# Automatic Estimation of Modulation Transfer Functions

Matthias Bauer[1,2], Valentin Volchkov[1], Michael Hirsch[3,†], Bernhard Schölkopf[1,3,†]

[1]Max Planck Institute for Intelligent Systems, Tübingen, Germany
[2]University of Cambridge, Cambridge, United Kingdom
[3]Amazon Research

`first.last@tue.mpg.de, hirsch@amazon.com`

## Abstract

*The modulation transfer function (MTF) is widely used to characterise the performance of optical systems. Measuring it is costly and it is thus rarely available for a given lens specimen. Fortunately, images recorded through an optical system contain ample information about its MTF, only that it is confounded with the statistics of the images. This work presents a method to estimate the MTF of camera lens systems directly from photographs, without the need for expensive equipment. We use a custom grid display to accurately measure the point response of lenses to acquire ground truth training data. We then use the same lenses to record natural images and employ a supervised learning approach using a convolutional neural network to estimate the MTF on small image patches, aggregating the information into MTF charts over the entire field of view. It generalises to unseen lenses and can be applied for single photographs, with the performance improving if multiple photographs are available.*

## 1. Introduction

Never before has photography been as widespread as today, thanks in part to the adoption of ever-improving, affordable digital sensors. Most cameras used today are owned by non-professional photograph enthusiasts whose access to high-quality lenses and expensive laboratory equipment is limited.

The quality of an optical system depends on all of its components and in particular on lens and sensor. Increased resolution and sensor quality implies a corresponding need to employ higher quality optics. While an ideal lens produces a perfect point response (modulo diffraction), real lenses are plagued by a plethora of optical aberrations, such as chromatic aberrations, coma, or field curvature inducing position-dependent defocus.

Various methods exist to characterise the properties and establish the quality of lenses, ranging from visual inspection of test photographs to the complete measurement of the distorted wave front with a Hartmann-Shack sensor [1]. In principle, all information about lens aberrations is captured by the *point spread function* (PSF). The PSF describes how a perfect point source is blurred, and it is spatially varying across the field of view. Direct measurement of the PSF is difficult and time-consuming. In practice, the *modulation transfer function* (MTF) is used as *de facto* standard quality measure of camera lenses [2]. It can be computed from the PSF and it encodes the frequency and direction dependent local relative contrast. Intuitively, it encodes how the contrast of a perfect sine grating is diminished by aberrations of an optical system, as a function of the width of the grating, cf. Fig. 2. Most commonly, the MTF is measured from photographs of standardised test chart [3], but more thorough techniques exist and professionals employ specialised MTF test stations, for example to adjust misaligned lenses [2].

All methods have in common that they require additional equipment and are, thus, not suitable for a large number of photography enthusiasts who wish to characterise their lenses, or to determine whether their equipment is up to its specifications. However, every photographer has access to a large collection of unprocessed high resolution photographs of natural scenes. While image statistics are scene dependent, average statistics over many images or patches tend to be universal [4–6]. Moreover, the success of blind image deblurring for optical aberration correction [7–10] lead Tang and Kutulakos [6] to conjecture that "single-photo aberration estimation and depth recovery may indeed be possible".

In this paper we present a method for automatic MTF estimation of a camera lens system directly from photographs taken with that system. We employ a convolutional neural network (CNN) that takes image patches as inputs and returns corresponding MTF values. A set of debayered but otherwise unprocessed RAW photographs is decomposed into patches and their local MTF estimates are aggregated into a single MTF chart over the entire field of view of the lens. Our architecture is flexible and can use both a single photograph as well as a set of photographs for improved performance by averaging patches in feature space. While the estimates are not as accurate as photometric measurements, they are much easier to perform, faster and characterise many qualitative and quantitative features of the MTF curves well. To gather ground truth training and validation data, we built a 2 m × 1.5 m pinhole display to accurately and efficiently measure the PSFs of a lens from a small number of images.

---

[†]The scientific idea and a preliminary version of code were developed prior to joining Amazon.

**Main contributions.**

1. We present, to the best of our knowledge for the first time, an algorithm for automatic estimation of the *modulation transfer function* (MTF) of a camera lens system directly from photographs taken with it. Our method is fast and does not require any additional equipment such as test charts or optical benches.

2. We built a new dataset of ground truth *point spread functions* (PSFs) of lens aberrations for several consumer lenses across the entire field of view, which is publicly available on the project website[1]. The PSFs were acquired using an extended pinhole setup for accurate and efficient measurement of the lens PSFs.

## 2. Background

### 2.1. Image formation model

While the *point spread function* (PSF) is typically non-stationary across the field of view, we assume that the PSF can be considered locally homogeneous across a small image patch. Similar to Schuler et al. [11] we assume the following global image formation model

$$y = \sum_i h_i * (w_i \odot x) + \epsilon \qquad (1)$$

where $x$ denotes the sharp and $y$ the blurred image, $w_i$ is a windowing function that cuts out the $i$th patch at location $(r_i, \varphi_i)$ from the sharp image, $h_i$ is the corresponding local PSF, and $\epsilon$ denotes additive Gaussian noise, i.e. $\epsilon \sim \mathcal{N}(0, \sigma^2)$. In a local patch with homogeneous blur, the image formation model simplifies to $y_i = h_i * x_i + \epsilon'$. Using a point source or pinhole as object directly (i.e. $x = \delta$) yields the PSF as image.

Ideal lenses are rotationally symmetric, i.e. the PSFs in the corners are rotated versions of each other. Real lenses consist of many elements that can be misaligned and are often surprisingly asymmetric [2]. Following Hirsch and Schölkopf [8], we introduce global polar coordinates $(r, \varphi)$ on the entire image and local Cartesian coordinates $(u, v)$ on patches; the local coordinate system is rotated according to the location of the patch such that $u$ denotes the radial and $v$ the tangential direction, respectively, see Fig. 1. Thus, the PSF becomes a function of both coordinate systems $\text{PSF}_\theta(u, v; r, \varphi)$, and $\theta$ denotes the camera and lens settings that influence the PSF, such as aperture and focus [6].

[1] https://ei.is.mpg.de/project/mtf-estimation

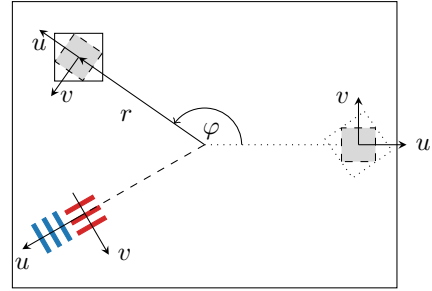

Figure 1. Global image coordinate system $(r, \varphi)$ and local patch coordinate system $(u, v)$. The latter is rotated to be aligned with the radial and tangential direction. *Sagittal lines* (■) can be used to measure the *MTF in the tangential* direction. Tangential MTF values are denoted by solid lines (≡) in MTF charts. *Meridional lines* (■) can be used to measure the *MTF in the radial* direction. MTF values are denoted by dashed lines ( ┊ ) in MTF charts.

### 2.2. Modulation transfer function (MTF)

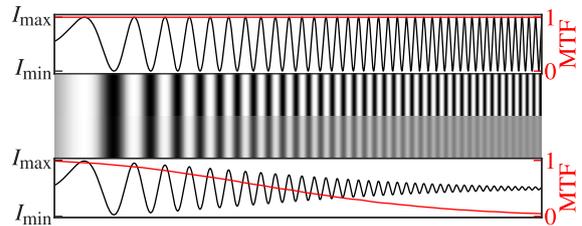

Figure 2. The *modulation transfer function* (MTF) characterises the diminished relative contrast (——) due to blur of a sine grating with varying frequency. *top:* the original grating has perfect contrast at all frequencies; *bottom:* the blurred pattern shows diminishing contrast at higher spatial frequencies (finer grating).

The *modulation transfer function* (MTF) is a function of spatial frequency and can be defined in two ways: 1) via photometry, and 2) via the Fourier transform of the PSF. In photometric terms, the MTF characterises the diminished contrast due to image blur at a particular spatial frequency: Once homogeneous blur is applied to a sine grating with perfect contrast at all frequencies (Fig. 2 *(top)*), the contrast diminishes, especially for higher spatial frequencies (Fig. 2 *(bottom)*). Formally, the MTF is defined as the relative contrast $C(f)$ between maximal and minimal intensity ($I_{\max}(f)$ and $I_{\min}(f)$) at a certain spatial frequency $f$, normalised by its zero frequency component $C(0)$:

$$\text{MTF}(f) = \frac{C(f)}{C(0)}; \quad C(f) = \frac{I_{\max}(f) - I_{\min}(f)}{I_{\max}(f) + I_{\min}(f)} \qquad (2)$$

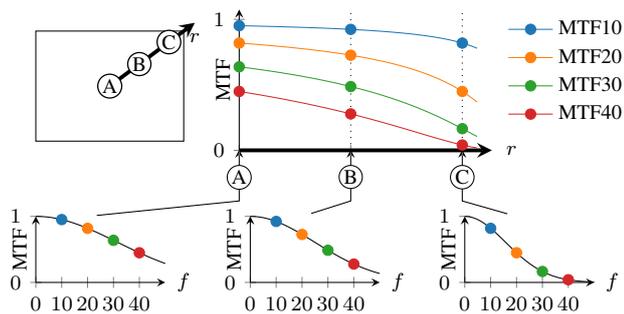

Figure 3. Several local MTF measurements (bottom, MTF vs. spatial frequency $f$) at positions Ⓐ, Ⓑ, Ⓒ are aggregated into a single global MTF chart (top; MTF10/20/30/40 vs. radial position $r$).

The MTF can also be related to the two-dimensional complex Fourier transform of the PSF, which is commonly referred to as *optical transfer function* (OTF). The OTF can be decomposed into its amplitude and phase, the MTF and the *phase transfer function* (PhTF), respectively:

$$\text{PSF}(u,v;r,\varphi) \xrightarrow{\mathcal{FT}} \text{OTF}(f_u,f_v;r,\varphi) \propto |\text{MTF}|e^{i\text{PhTF}}, \tag{3}$$

where both MTF and PhTF depend on the spatial frequencies $(f_u, f_v)$ and the patch location $(r, \varphi)$.

Lenses are often characterised by the MTF values at certain spatial frequencies, namely at $10, 20, 30, 40\,\text{cycles/mm}$ [2]; the associated MTF values are denoted as MTF10/20/30/40[2]. Due to azimuthal symmetry of the lens, one typically distinguishes between MTF in radial and tangential direction. Fig. 1 shows meridional (concentric) lines to measure the MTF in radial direction $u$, whereas the MTF in the tangential direction $v$ is measured using sagittal (radial) lines. To summarise the radial and tangential MTF over the field of view of a lens, one often plots the values for MTF10/20/30/40 from the image centre ($r = 0$) to the corners ($r \approx 21.63\,\text{cm}$), cf. Fig. 3. We refer to this summary plot as *global MTF chart* and note that this type of plot is typically provided in datasheets by lens manufacturers [12].

## 3. Related Work

Our work on blind MTF estimation from real images lies at the intersection of traditional photometric measurements and computer vision.

**Photometric MTF measurement.** There exists a range of methods to measure the MTF of an optical system. Perhaps most widely used and implemented in commercial software is the slanted-edge method [3, 13], which measures the MTF perpendicular to a perfect edge and relies on the Fourier transformation of the *lines spread function* (PSF integrated along one dimension) derived from that edge. Several extensions and alternatives have been proposed that employ other or extended patterns, e.g. [14–17]. All methods require standardised test charts and typically only yield a very small number of MTF measurements over the field of view. Moreover, careful calibration and lighting is necessary to obtain reliable and consistent results.

Our *PSF panel* belongs to a different class of methods, which use point light sources, also referred to as pinholes or artificial stars, to directly measure the full PSF, from which the MTF can be computed [18, 19]. A perfect lens would map one point source onto a single pixel. Navas-Moya et al. [20] use an LCD screen to simulate pinholes; however, they are limited by the resolution (minimal pinhole size) and size of the screen, which preclude MTF measurements for wide angle lenses and high-resolution DSLR cameras. Our screen is substantially larger and has pinholes of finer diameter.

All above methods, including ours, do not measure the MTF of the lens but rather the MTF of the combined camera lens system. Commercial MTF test stations as used in [2] do not suffer from this shortcoming but are more expensive.

**MTF estimation from natural images.** There exists little prior work on automatic MTF estimation from natural images. Several methods limited to aerial photography rely on hand-crafted features (image variograms) to estimate parameters of a single homogeneous Gaussian MTF [21, 22].

**Blind image deblurring and PSF estimation.** While our method does not perform image restoration, MTF estimation is closely related to blind image deblurring and PSF estimation.

[23–25] estimate the PSF from known test patterns to perform non-blind deconvolution on blurred photographs. [23] also perform blind PSF estimation for unimodal blurs by predicting sharp edge locations. [26] estimate the PSF from a single image by optimising the parameters of the lens prescription model that is used to simulate the PSF similar to the optics software Zemax. Their method is fast and has few parameters but requires a lens prescription model. [11] propose a method for automated PSF capture with a single pinhole source and devise a non-blind correction method for optical aberrations. In [7] they extend their method to the blind case by making symmetry assumptions about the unknown PSF. [8] compare state-of-the-art blind deconvolution methods for lens blur, such as [10], and propose a PSF estimation based on kernel regression that allows the integration of multiple images. [9] use visual and geometric priors to perform aberration correction and assume that the PSF is rotationally symmetric. [2] report this assumption to be true for less than 10% of real lenses. [27] present a parametrized model of spatially varying optical blurs, and [6] provide a theoretic analysis of image formation under Seidel aberrations and its

---

[2] Not to be confused with the common notation MTF50, which is used to denote the spatial frequency at which the MTF attains the value 0.5

consequences on blind and non-blind PSF estimation and depth-estimation. They conclude that "single-photo aberration estimation and depth recovery may indeed be possible". Our work addresses the former domain and focuses on planar scenes that are approximately within the focus plane to avoid additional blur due to defocus.

## 4. PSF Measurements using PSF Panel

In this section we briefly present our *PSF panel* to obtain ground truth PSFs for lens aberrations. For details of the specifications and the data acquisition, see Supplement. Fig. 5 shows several ground truth PSFs as well as the global MTF charts along one diagonal of the field of view for three lenses, which have been recorded with the *PSF panel*. For further measurements on a range of other lenses, see Supplement.

### 4.1. Specifications of the Panel

The *PSF panel* consists of an LED-backlit glass plate and a photographic film attached to the glass. The LED-panel is $2\,\text{m} \times 1.5\,\text{m}$ in size and emits white light (6500 K) homogeneously distributed over its area. The point sources for the PSF measurements are implemented by covering the LED-panel with a black photographic film with a pattern of transparent dots. The diameter of the dots is $150\,\mu\text{m}$, and the dots are arranged in a square grid at a distance of $25\,\text{mm}$, yielding in total $80 \times 60$ pinholes, see Fig. 4.

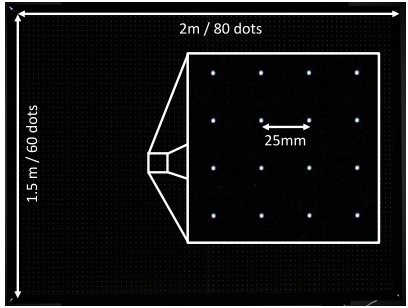

Figure 4. Photograph of the *PSF panel* used to obtain ground truth PSFs. The inset shows a detail of the pinholes.

We recorded the PSFs of several consumer DSLR lenses using a 50.6 MP Canon EOS 5DS R camera body. Images of the PSF-panel were taken at that distance where 80 pinholes filled up the horizontal direction, leaving 53 pinholes in the vertical direction. The distance between the PSFs on the images therefore amounts to 111 pixels. Altogether, the size of the pinholes in the image-plane is smaller than one pixel; hence the pinholes act as a point source.

### 4.2. Data collection, processing, and PSF extraction

Fully automated data acquisition was performed using a bash-script running a series of `gphoto2` commands. Prior to the actual acquisition the lens was focused using a Python script that minimizes the recorded image size of central pinholes. Due to chromatic aberrations, not all colour channels are in focus at the same time. In this work, we used the green colour channel for focus adjustment and restrict our further analysis to image data from the green colour channel only.

The data set was recorded with four aperture settings for each lens: open aperture, 2.8, 4, and 5.6. Especially in the case of open aperture the size of the PSF can vary substantially between the centre and the edges of the image. Therefore, a series of different exposure times was taken for each aperture setting and results were averaged over 10 images in order to improve the signal to noise ratio.

All images of the PSF-panel were taken in the camera RAW format and subsequently developed using `dcraw`. The developed images were averaged and the positions of the PSF-centroids were extracted. Out of all exposure times, the PSFs with the longest exposure times but without saturated pixel values within a $111 \times 111$ pixel patch were selected. The background level of each patch was determined from its 4 corners and used to threshold and segment out the PSF.

### 4.3. Fast kernel regression

To further reduce noise and to interpolate PSFs at unobserved locations, we developed a sped-up version of the kernel regression by Hirsch and Schölkopf [8] that interpolates the PSF $h(\mathbf{x}), \mathbf{x} = (u, v; r, \varphi)$, at a new location $(r, \varphi)$:

$$h(\mathbf{x}) = \frac{\sum_i h_i(\mathbf{x}_i) K(\mathbf{x} - \mathbf{x}_i)}{\sum_i K(\mathbf{x} - \mathbf{x}_i)} \quad (4)$$

where $K(\cdot)$ is a squared exponential kernel that factorises over its dimensions and has lengthscale $\ell_x$ for each factor $x \in \mathcal{X} = \{r, \varphi, u, v\}$:

$$K(\mathbf{x}) = \prod_{x \in \mathcal{X}} K_x(x), \quad K_x(x) = \exp\left(-\frac{\|x\|^2}{2\ell_x^2}\right) \quad (5)$$

The index $i$ in Eq. (4) runs over all pixels of all recorded PSFs: To obtain the value of one pixel at local coordinates $(u, v)$ for one PSF at a new location $(r, \varphi)$ the original algorithm computes the covariance with 4800 PSFs with $111 \times 111$ pixels $\approx 6 \cdot 10^8$ data points, which is clearly infeasible. We can exploit the product structure of the kernel and the discrete nature of the $(u, v)$ coordinates to dramatically speed up this computation, see Supplement for details. We found the associated approximation errors to be negligible in practice.

## 5. Estimating the MTF from Photographs

We aim to build a learning system that produces the global MTF charts for a lens, given a photograph or a set of photographs captured with that lens. As the MTF is sensitive, e.g. to JPEG Compression [28] or Gamma Correction [29], high resolution and unprocessed images are required, see Supplement for capture and RAW development details.

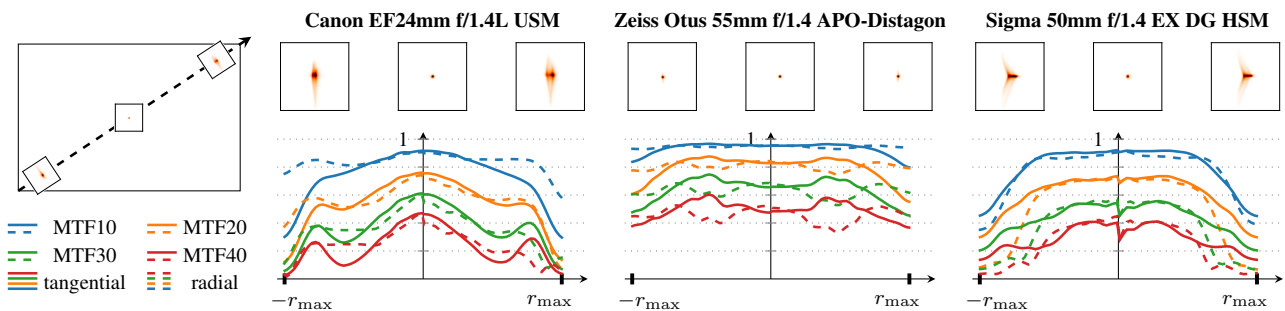

Figure 5. Ground truth global MTF charts in radial (┆┆┆) and tangential (≡≡≡) direction along the diagonal from the bottom left to the top right (- - ➤) as measured with our *PSF panel*. PSFs on $111 \times 111$ patches (top) are extracted at $-r_{\max}, 0, r_{\max}$ and rotated into the common coordinate frame. F.l.t.r: Canon EF24mm f/1.4L USM, Zeiss Otus 55mm f/1.4 APO-Distagon, Sigma 50mm f/1.4 EX DG HSM (all at f/1.4).

Similar to approaches in blind image deblurring with inhomogeneous blur, e.g. [10, 30], we first estimate the MTF values on small patches, over which the blur is assumed to be uniform. In a second step these local estimates are aggregated into a consistent global MTF chart. For local estimation we employ a deep convolutional neural network that performs regression from the input patches to the final MTF values. To obtain a globally consistent MTF model we use Gaussian Process regression [31] to smoothen and interpolate the noisy estimates in a non-parametric way.

Critically, our system should be able to fuse information from several photographs, which have all been captured with the same camera lens system. The PSF, and consequently the MTF, only depends on the system and its settings; and while it varies across the field of view, it is the same for patches extracted from the same location in different photographs, e.g., always the top right corner. These patches all have different image content but share a common blur kernel.

As lens PSFs can differ between colour channels, we treat each channel separately. For simplicity, we restrict ourselves to the green channel in this work.

### 5.1. Network architecture for local MTF estimation

The network is composed of three main components, see Fig. 6: (i) initial data processing, (ii) a convolutional network (CNN) that produces an intermediate feature representation of the input, and (iii) three fully connected layers (FC) to perform regression onto the output.

**Inputs.** The network uses single image patches of size $192 \times 192$ as input. To account for the rotating local coordinate system, the patches have been rotated by an angle $-\varphi$, such that the radial and tangential direction (local coordinate system $(u, v)$) are aligned with the horizontal and vertical axes, see Fig. 1; we have to extract correspondingly larger patches from the original image to allow for this rotation.

Thus, the network always has to predict the MTF values in horizontal and vertical direction only. The maximal range of the input values is scaled to $[0, 1]$, and the blurred images are subsequently mean normalised.

**Initial data processing.** It is possible to train the network to predict MTF values both in radial and tangential direction simultaneously. However, to simplify the task, we only predict the MTF in the horizontal direction $(u)$. In order to also obtain the MTF in the orthogonal $(v)$ direction, we flip a copy of the input patch by $-90°$ and independently pass it through the same network. The predictions for both copies are then concatenated. This procedure not only aids learning but helps the network to generalise better as it becomes more robust against correlations of the MTF values that might be present in training but not in test data. As edges are discriminative features for blind PSF estimation [23, 32], we append the gradient (Sobel filtered) image along the direction, in which we estimate the MTF, as a second input channel[3]. Further, we subsample the spatial dimensions of the input into channels to allow early convolutional layers to access a larger field of view [33]. That is, we subdivide the input patches into non-overlapping groups of $M \times M$ pixels and move every pixel to one of $M^2$ channels depending on its location in the $M \times M$ group, see Supplement for details. Each channel then corresponds to a subsampled and slightly shifted version of the input.

**CNN and FC.** The CNN consists of an initial convolutional layer followed by a series of residual blocks [34] that use strided convolutions to reduce the spatial dimension of the input and increase the feature size. The resulting feature representation is then passed into a series of fully connected layers that regress onto the MTF outputs. All activation functions are ReLUs [35] except for the last layer which uses a sigmoid activation, as MTF values lie between 0 and 1. We use an $L_2$ (squared error) loss function between the predicted and the true MTF values. The network is trained end-to-end using the Adam optimiser [36] with a decaying learning rate schedule. For details, see Supplement.

---
[3]In principle, the network could also learn a gradient filter but we found that adding it manually improved performance.

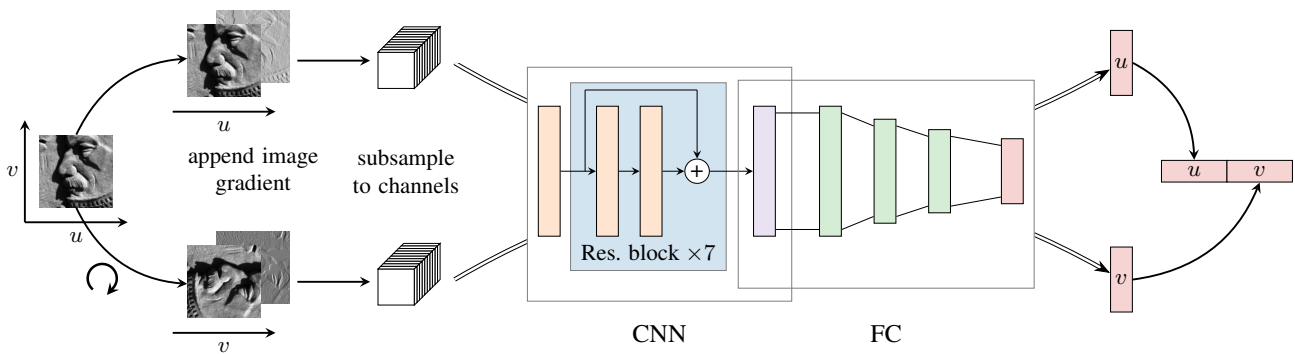

Figure 6. Data processing and network architecture. The input image is duplicated and flipped, the image gradient in the horizontal direction is appended, and the spatial dimensions are subsampled into channels. Both tensors are passed independently through the same network, which predicts the MTF values in the horizontal direction. Both outputs are concatenated to yield radial and tangential MTF values. The network consist of a CNN with initial convolution, 7 residual blocks, and three fully connected layers. When using multiple input images, the intermediate feature representation of the inputs are computed separately and then averaged before being passed through the FC.

**Outputs.** The frequency axis of the local MTF plots depends on the pixel pitch of the sensor. To make the network independent of the sensor pitch, the actual outputs are MTF values at fractions of the Nyquist frequency measured in pixels; the appropriate MTF values for 10, 20, 30, and $40\,\text{cy/mm}$ are then interpolated according to the particular pixel pitch of the camera. For our Canon EOS 5DS R the pixel pitch is $4.14\,\mu\text{m}$ leading to $f_{\text{Nyquist}} = 120.7\,\text{cy/mm}$.

**Multiple input patches.** We use a simple extension of the above network to deal with multiple input patches that have been blurred by the same PSF and, thus, have identical MTF values, for example for patches from the same location $(r, \varphi)$ but from different images captured with the same lens.

Multiple patches are treated as follows: (i) we compute the intermediate feature representation (feature activation of the last convolutional layer; purple in Fig. 6) for each patch individually; (ii) we average these feature activations elementwise; and (iii) we feed the averaged activation through the fully connected network to obtain a single prediction. We pre-trained the network on individual patches and used four patches with identical MTF values during training. At test time we can compute feature activations for more or less than four patches and average them in the same way, making our method flexible and agnostic to the order of patches [4].

---

[4] Alternatively, one could concatenate the feature representations and pass them into the FC layers as a long vector. However, this approach would have three immediate drawbacks: (i) the size of the FC layers would need to be substantially larger (ii) the order of the patches would matter (iii) we would be strictly limited to the same number of patches at test time.

## 5.2. Supervised training procedure and datasets

To train the network for local MTF estimation we construct the following supervised learning task: We synthetically blur sharp image patches with PSFs for which we analytically compute the MTF values; these blurred patches are then used as inputs for the network to predict the corresponding MTF values as labels. We draw random combinations of image patches and PSFs, such that the network never sees the same training example twice. In the following we briefly describe the datasets of sharp image patches and ground truth PSFs. For further details, see Supplement.

**Sharp image patches.** We use two different sources of sharp images, see Fig. 7: (i) a regular checkerboard-like pattern with edges in all directions, which was proposed by Joshi et al. [23] for PSF estimation; (ii) patches from real photographs of natural scenes that have been captured with a sharp high-end lens (Zeiss Otus 55mm f/1.4 APO-Distagon), small aperture and under good light conditions using a tripod. For the regular pattern we use random sizes, rotations and contrast to simulate varying conditions at test time, for which lighting, orientation and scale depend on the lens, size of printout, and distance to the printout. Sharp patches from real photographs are extracted randomly from the central region of the images[5] and downsampled by a factor of two to further reduce the effective PSF. As the statistics between natural images and the regular pattern are significantly different, we train separate networks for both sources.

**Ground truth PSFs/MTFs.** We use two different types of PSFs: (i) real PSFs recorded with our PSF panel, and (ii) artificially generated PSFs: a sum of two Gaussian blurs – a narrow central peak and a wider wing – of varying widths along the principle axes and with large eccentricity. We found that adding the artificial PSFs improved performance, probably due to larger variability of shapes in the combined dataset.

---

[5] A rectangular region with half the image dimensions

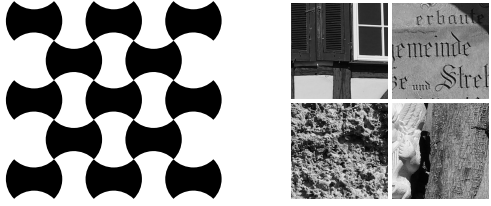

Figure 7. Two sources of "sharp" input data. *left:* Regular pattern proposed by Joshi et al. [23]; *right:* patches from real photographs.

## 6. Experiments

In the following, we present results for MTF estimation from lens-blurred photographs of the regular test pattern and natural scenes. The results take the form of MTF charts for tangential and radial MTF10/20/30/40 values, see Fig. 3. The MTF charts are either for a fixed angle along a ray from the centre to the top right corner, or averaged over the angular coordinate ("azimuthal average"). We present results for the Sigma 50mm f/1.4 EX DG HSM, whose PSFs have not been used during training. Results for other lenses and further experimental details can be found in the Supplement.

### 6.1. MTF estimation on the regular pattern

To start, we consider the regular test pattern in Fig. 7 *(left)*. Similar to other methods using test charts, our aim is to estimate the MTF from photographs of the pattern. Contrary to other methods, we do not explicitly use correspondences between the ground truth test pattern and the photographs other than training the network solely on synthetically blurred patches from the pattern. Fig. 8 shows MTF charts for the Sigma 50mm from the centre to the top right corner.

First, we check the performance of our network on synthetically blurred patterns that are prepared in the same way as the training data (Fig. 8 *(right)*) and compare them to the ground truth measurements (Fig. 8 *(middle)*). The estimates agree almost perfectly to the ground truth, indicating that the network has learned its task well; we found the overall error averaged over the azimuthal direction to be smaller than 5%.

To inspect the behaviour on real photographs, we use a printout of the test pattern of size A1[6] and take photographs to cover the entire camera screen or just one of the four quadrants, see Supplement. For a given location $(r, \varphi)$ in the image, we extract patches from the two photographs at locations $(r, \varphi + \Delta_i), \Delta_i \in \{-0.02, 0, 0.02\}$ to increase the number of patches that have been blurred with approximately the same PSF from 2 to 6. We inspect the influence of the number of patches for MTF estimation on natural scenes in Sec. 6.2. Fig. 8 *(left)* shows the obtained estimates from the centre to the top right corner of the image. They agree well both qualitatively and quantitatively with the ground truth measurements (Fig. 8 *(middle)*), though the radial MTF falls off more quickly in this case. In Fig. 9 we show results averaged over the angular coordinate instead of just a slice from the centre to the top right corner. They are also in good agreement to the ground truth data and the averaged absolute error is smaller than 0.1 MTF units (10% of the maximum) in this case and generally smaller than 0.15 MTF units for other lenses, see Supplement. We observe a small bias to under-estimate the MTF values, which we attribute to the MTF of the printer used to print the test pattern.

### 6.2. MTF estimation on natural scenes

We now turn to the estimation of MTF values from photographs of natural scenes, see Supplement for examples. We use the same architecture as for the regular test pattern but train the model on synthetically blurred patches from photographs of natural scenes as explained above, see Fig. 7 *(right)*. Fig. 10 shows results for a fixed angle along a ray from the centre to the top right corner (top row), as well as averaged over all angles (bottom row). The estimates are more noisy than for the regular pattern; to obtain a smooth and globally consistent MTF chart, we fit a Gaussian Process (GP) regression to each MTF frequency and direction. For the azimuthal averages, we use the values from the GP mean.

While the shape of the estimated curves agrees well with the ground truth, the absolute values are larger and the curves do not fall off as much towards the corners. We observed this over-estimation consistently for all lenses and explain it as follows: While perfectly sharp patches were used to train on the regular pattern, we used patches from actual photographs in this case. Even the sharpest lens and downscaling still leave a small blur on the "sharp" training patches, which were, thus, effectively blurred twice: once when collecting the "sharp" images and once synthetically. At test time, the images are only blurred once by the lens used to capture the photo. We compensate for this effect by multiplying our estimates by the effective MTF of the downsampled PSF of the Zeiss Otus lens used to capture the "sharp" training images. We estimate these compensation factors as 0.98, 0.95, 0.9, 0.83 for MTF10, MTF20, MTF30, and MTF40[7] and employ them in all results except for Fig. 10 *(top)*, which shows raw values. This compensation improves the results for all lenses, and we found the average error for the Sigma 50mm to be smaller than 0.1 MTF units in the centre and smaller than 0.2 MTF units towards the corners. For other lenses, the average error was between 0.1 and 0.2 MTF units, see Supplement.

---

[6]pattern period: 25 mm, we glue four A3 printouts together, making it feasible for non-professionals

[7]We use constant factors as the MTF values of the Zeiss Otus only change very little towards the corners for small apertures

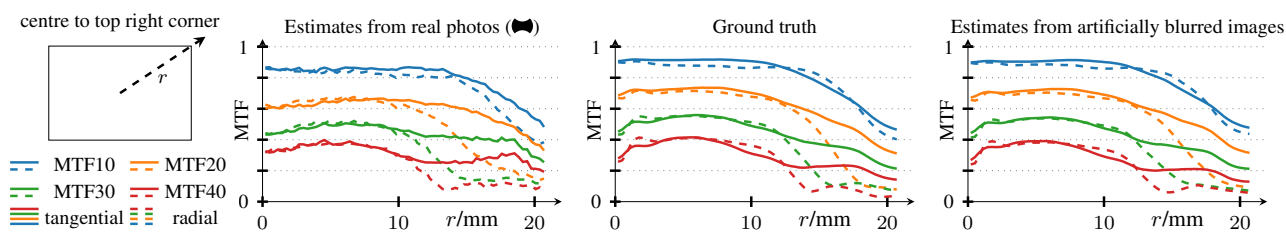

Figure 8. Results of MTF estimation for the regular pattern ⋈ in Fig. 7 *(left)* for the Sigma 50mm f/1.4 EX DG HS lens at f/1.4. Patches are extracted along the diagonal from the centre of the image to the top right. *left:* Estimates from photographs of a printout of the pattern; *middle:* Ground truth obtained from kernel regressed PSFs at the same locations as the patches in *(left)*; *right:* Estimates from sharp training patches that where artificially blurred using the ground truth PSFs from *(middle)*. For more results, see Supplement.

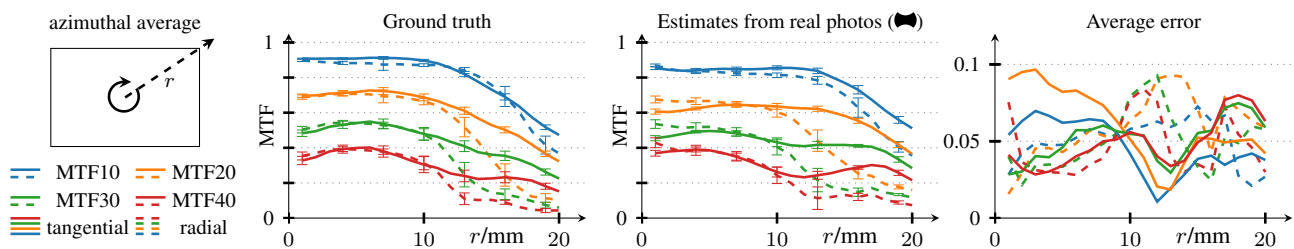

Figure 9. Results of MTF estimation for the regular pattern ⋈ for the Sigma 50mm f/1.4 EX DG HS lens at f/1.4. Results are averaged over the angular variable. *left:* Ground truth obtained from kernel regressed PSFs; *middle:* Estimates from photographs of a printout of the pattern; *right:* Average error between estimates and ground truth. For more results, see Supplement.

In Fig. 11 we explore how the estimates depend on the number of patches from different photographs used for local MTF estimation at a single location $(r, \varphi)$. While the general scale can be identified from a single photograph, the qualitative shape of the curve can not. Moreover, the individual estimates are very noisy (large Gaussian Process variance). Upon inclusion of more patches from the same location $(r, \varphi)$ but from other photographs, the quality of the predicted MTF curves improves, and the variance of the estimates decreases.

### 6.3. Discussion of discrepancies and limitations

We partly explain the discrepancy between our estimates and the ground truth with the field curvature (curved focus plane) of camera lenses: While the PSF panel is perfectly flat, natural scenes typically have a continuous range of depth values. Thus, some object parts in the image corners are likely to lie in the curved focus plane, appearing sharper than the measured PSFs on our flat panel. Non-planar scenes can also lead to underestimation of the MTF if parts of the scene are not in focus. Other failure cases include large textureless surfaces such as sky, which do not contain information about the MTF, or patches that only contain edges in one direction, see Supplement for an example. We mitigate these effects by selecting as planar and textured natural scenes as possible. In a deployed system a pre-filtering step would be necessary to identify "good" patches; for example, Hu and Yang [32] propose a method to find good regions for PSF estimation.

### 6.4. Comparison to other methods

To the best of our knowledge, we present the first general method that is capable of estimating lens MTF charts from natural photographs. We compare our method to (i) PSF estimates obtained with blind deblurring algorithms, from which we compute the MTF. (ii) photometric measurements using MTF test charts

**Comparison to blind image deblurring.** We compare to the state-of-the-art single image blind deblurring algorithm by Michaeli and Irani [10], which provides estimates of the corresponding local PSFs. We use a single photograph of a natural scene to estimate the PSFs, and then compute the corresponding MTFs. The algorithm requires relatively large patches and is very slow (several hours *for a single patch* of size $800 \times 800$). Our method estimates the *entire MTF chart* for the same photograph in minutes. As blind deblurring uses large patches, we perform multipatch estimation on six co-located smaller patches with our method. Fig. 12 shows results averaged over all angles; our method is better able to estimate the MTF values, though it fails to capture the fall-off towards the corners accurately. The PSF estimates from blind deblurring are larger than the ground truth PSFs and lead to severe underestimation of the MTF. We stress that our method improves in performance when using more than one photograph, see Figs. 10 and 11.

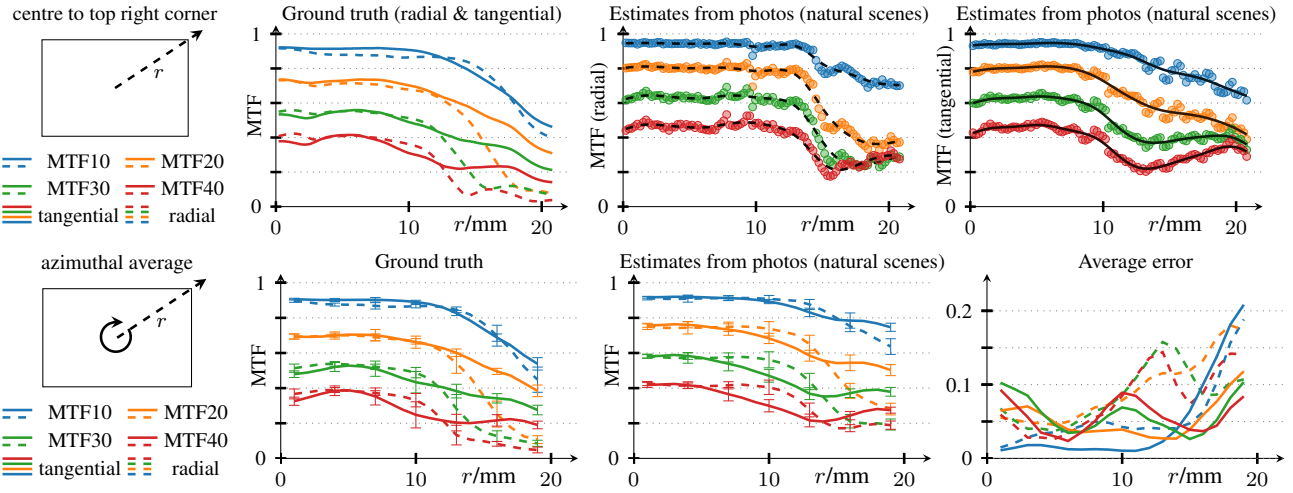

Figure 10. MTF estimation from photographs of natural scenes for Sigma 50mm f/1.4 EX DG HS lens at f/1.4. *top:* Estimates along a ray from the centre to the top right corner. The black lines indicate predictive mean functions from Gaussian Process regression. *bottom:* Estimates averaged over all angles. Errorbars indicate the variability of MTF in different directions. We compensated the average MTF values for the effective MTF of the Zeiss Otus used to capture the training photographs, see main text. For results on other lenses, see Supplement.

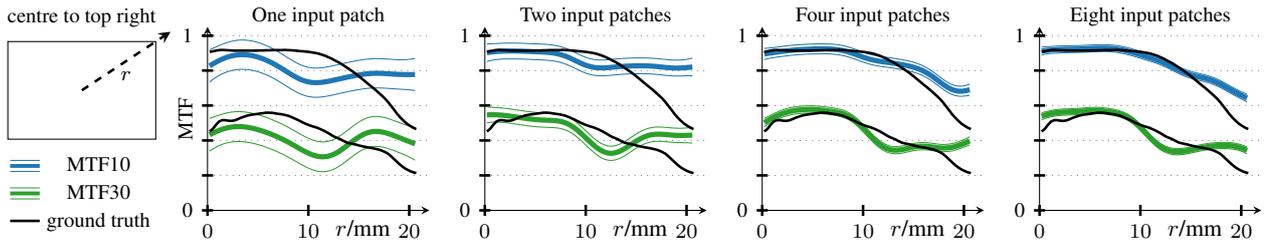

Figure 11. MTF estimates of MTF10 and MTF30 in the tangential direction for varying number of input patches from photographs of natural scenes. For each location $r$ along a ray from the centre to the top right corner we extract patches from 1, 2, 4, or 8 different photographs for multi-patch estimation. Results are interpolated with a Gaussian Process and we plot its mean and standard deviation.

**Photometric measurements.** We use a commercial MTF chart and the evaluation software iQ-Analyzer. The test chart contains 25 Siemens stars [14] and several slanted edges [13]. The stars are subdivided into eight segments on which the MTF is measured, providing MTF values at only 25 locations over the entire field of view. For estimates of radial and tangential MTF values we use for each star the segments corresponding to these directions. Slanted edges only provide the MTF in horizontal direction at only four locations (for this chart). We present results averaged over the angular variable in Fig. 13. The methods generally agree well qualitatively, i.e. in terms of the shape of the curves, but the estimates obtained with iQ-Analyzer are slightly lower.

## 7. Conclusion

We have presented a method for automatic estimation of the *modulation transfer function* (MTF) of camera lens systems directly from photographs of natural scenes captured using those systems. We envisage this method to be especially useful to users who wish to characterise their lenses without access to professional and expensive MTF measurement equipment and expertise in optical testing.

The contributions of the present paper are twofold: (1) We developed a novel method where, initially, sets of photographs are decomposed into patches, which are then processed by a trained convolutional neural network to estimate the local MTF. These local estimates are subsequently aggregated into a consistent global MTF chart using Gaussian Process regression. (2) We built a new dataset that enables statistical learning for our setup. It contains ground truth *point spread functions* (PSFs) of lens aberrations for several consumer lenses across the entire field of view and is publicly available on the project website[8]. The PSFs were acquired using an extended pinhole setup for accurate and efficient measurement of the lens PSFs from a small number of photographs. We use the dataset to artificially blur sharp images for training, as well as for the validation of the MTF estimates from photographs.

---

[8] https://ei.is.mpg.de/project/mtf-estimation

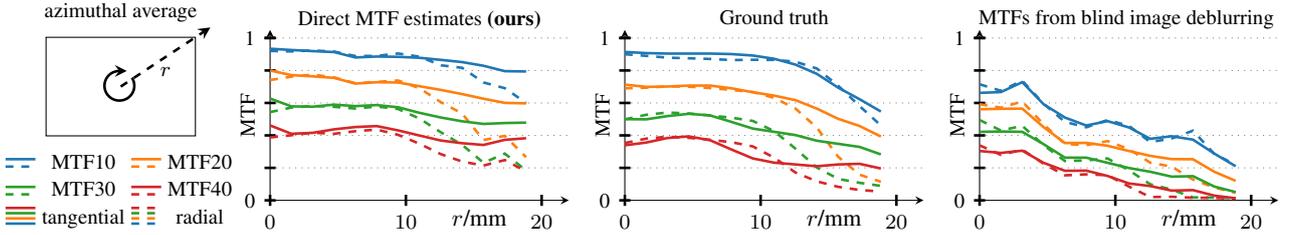

Figure 12. Comparison to state of the art blind image deblurring by Michaeli and Irani [10] on a photograph captured with the Sigma 50mm f/1.4 EX DG HS lens. *left:* MTF estimation using our method from a single photograph using six close-by patches for each estimate. *centre:* Ground truth obtained analytically from PSF measurements with our PSF panel. *right:* MTF estimates obtained analytically from PSF estimates with blind image deblurring [10].

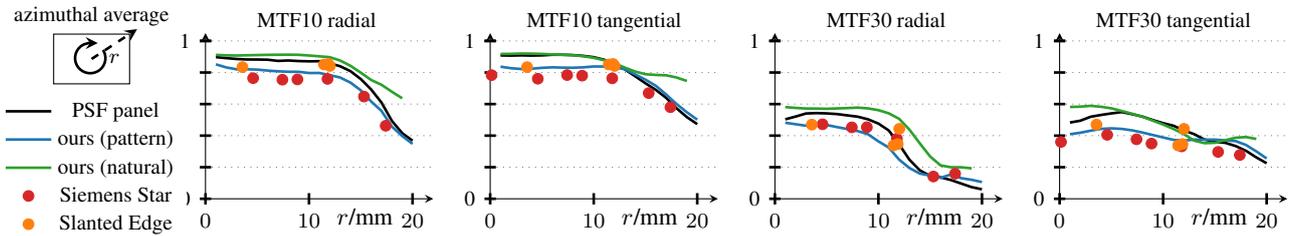

Figure 13. Comparison to photometric measurements from a commercial test chart with iQ-Analyzer for MTF10 and MTF30 in radial and tangential direction for the Sigma 50mm f/1.4 EX DG HS lens. MTF values are averaged over the angular variable for: the ground truth recorded with our PSF panel, estimates from photos on the regular pattern and natural images, and the photometric measurement from Siemens Stars and Slanted Edges. Values for Slanted Edge are only in the horizontal direction.

The resulting trained architecture is flexible and can handle both a single photograph as well as a set of photographs by automatically averaging patches in a feature space learned by the network. In the experimental validation, we estimated lens MTFs from (i) photographs of a regular checker-board like test pattern, as well as (ii) photographs of natural scenes captured in the wild. Our system is easy to use, and yields MTF estimates across the entire field of view of the lens within a few minutes. The estimates are in very good agreement with ground truth photometric measurements in terms of qualitative features of the MTF charts, and also yield reasonable quantitative performance. We outperform a baseline derived from blur kernel estimation with a state-of-the-art blind image deblurring algorithm.

Possible extensions in future work include (i) automatic pre-filtering and selection of patches, from which the MTF is computed, (ii) inclusion of and information fusion between different colour channels, and (iii) more sophisticated information fusion from several patches that goes beyond averaging in feature space.

## Acknowledgements

We thank the anonymous reviewers for their comments, Eduardo Pérez-Pellitero for help and feedback when writing the manuscript, Patrick Wieschollek and Mehdi Sajjadi for helpful discussions, and Alexandra Geßner for help with data acquisition. We thank Department Bülthoff at the MPI for Biological Cybernetics for access to their lab spaces. M.B. acknowledges partial funding by a Qualcomm Studentship and the EPSRC.

# Supplemental Material to "Automatic Estimation of Modulation Transfer Functions"

Matthias Bauer[1,2], Valentin Volchkov[1], Michael Hirsch[3,†], Bernhard Schölkopf[1,3,†]

[1]Max Planck Institute for Intelligent Systems, Tübingen, Germany
[2]University of Cambridge, Cambridge, United Kingdom
[3]Amazon Research

first.last@tue.mpg.de, hirsch@amazon.com

## Contents

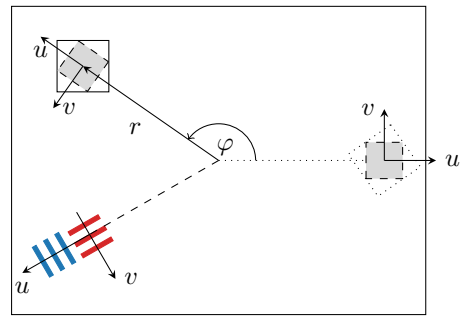

Figure 1. Global and local coordinate systems. A patch is extracted at global location $(r, \varphi)$ and rotated by an angle $-\varphi$ into a standardised common coordinate frame, in which the local patch coordinates in the radial and tangential direction $(u, v)$ are aligned with the image axes. *Sagittal lines* (■) run along the radial direction and can be used to measure the *MTF in the tangential* direction. They are denoted by solid lines (≡) in the MTF charts. *Meridional lines* (■) run along the azimuthal direction (along concentric circles) and can be used to measure the *MTF in the radial* direction, which is also referred to as *sagittal* direction. They are denoted by dashed lines (⋮⋮⋮) in our MTF charts.

## 1. Directional conventions

The nomenclature of MTF curves can sometimes be confusing. However, across science and industry, a consistent line style is used that we also adopted in our paper, see Fig. 1.

The confusion can arise from the distinction between the direction, in which the MTF is measured, and the orientation of a hypothetical stripe pattern, which is used to measure the MTF, see Fig. 1. The radial direction is also referred to as *sagittal* direction; thus, radial lines are also referred to as sagittal lines. However, these lines are then used to measure the MTF in tangential direction, which is also referred to as *meridional* direction.

In this work we use the terms *radial* and *tangential* MTF, always referring to the direction in which the MTF is measured.

## 2. Details of the PSF Panel

In this section we present our *PSF panel* to obtain ground truth PSFs for lens aberrations in more detail. Figs. 2 and 3 show a detail of the panel as well the general measurement setup that we elaborate on in the following subsections.

## 2.1. Specifications of the Panel

The PSF panel is $2\,\mathrm{m}$ by $1.5\,\mathrm{m}$ in size.[1] It consists of an acrylic glass plate with the height of $14\,\mathrm{mm}$, illuminated from two sides by opposing LED stripes, each containing 313 white LEDs ($6500\,\mathrm{K}$). The light is spread across the area of the panel owing to a system of light-guides within the glass. A diffusing screen with the height of $4\,\mathrm{mm}$ on top of the acrylic glass leads to a homogeneous light distribution. The point sources for the PSF measurements were obtained by covering the LED-panel with a black photographic film with a pattern of transparent dots. The film was attached to the diffusing screen using double-sided adhesive tape. The diameter of the dots was $150\,\mu\mathrm{m}$, the dots were arranged in a square grid at a distance of $25\,\mathrm{mm}$, yielding in total $80 \times 60$ pinholes, see Fig. 2.

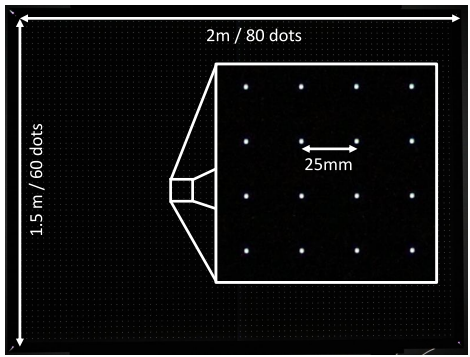

Figure 2. Photograph of the "PSF panel" used to obtain ground truth PSFs. The inset shows a detail of the pinholes.

We recorded the PSF of several commercial DSLR lenses using a $50.6\,\mathrm{MP}$ Canon EOS 5DS R camera body. The sensor has a resolution of $8688 \times 5792$ and is equipped with a filter that cancels the optical low-pass filter. Images of the PSF-panel were taken at a distance such that 80 pinholes filled up the horizontal direction, leaving 53 pinholes in the vertical direction. The distance between the PSF's on the images therefore amounted to 111 pixels. Altogether, the size of pinhole in the image-plane was smaller than one pixel[2], hence the pinholes acted as a point source.

## 2.2. Alignment of the camera

We geometrically aligned the camera with respect to the PSF-panel by eliminating the degrees of freedom one by one whenever it was possible. We used the live mode of the camera as well as external tools to position the camera. First, we adjusted the orientation of the PSF-panel using an electronic bubble level with a leveling accuracy of $\pm 0.2°$. Therefore,

---

[1] FLM High Power, manufactured by Lichtraum, www.lichtraum-muenchen.de
[2] $25\,\mathrm{mm}/111\,\mathrm{pixel} = 225\,\mu\mathrm{m} > 150\,\mu\mathrm{m}$

the face of the panel was aligned parallel to gravity, while the long side was aligned orthogonal to the gravity. Then, we mounted the camera on a stable tripod with a geared column and a geared head. The height and the vertical angle were aligned using an iterative procedure. As step one, we positioned the camera as close to the PSF-panel as possible and used the geared column to get the image of the center of the panel onto the center of the sensor. In the second step, we moved the camera several meters away from the panel and used the vertical angle adjuster to again center the panel on the sensor. These two steps were iterated until no adjustments were required and the panel was in the center of the image at any distance. The working distance was chosen such that the image contained 80 pinholes in the horizontal direction. We monitored and fixed that distance using a laser distance meter attached to the hot-shoe of the camera.

In order to decouple the horizontal angle and the horizontal position we set up a vertical laser-line, as shown in Fig.3. The laser-line originated from the center of the upper side of the panel and propagated perpendicularly with respect to the panel. For that we installed a laser pointer on the upper left corner of the panel (see *top view* in Fig.3). The laser beam was carefully aligned along the upper side, then a $90°$ prism was introduced into the beams path in the center of the upper side. Fine adjustment of the $90°$ reflected beam was then achieved by making sure that the weak back-reflection from the entering surface of prism went exactly back into the laser. A diffractive optical element directly at the exit of the prism was used to create a vertical line in the semi-space in front of the panel. Thus, this line marked the horizontal center of the panel. The camera was then positioned horizontally such that the laser line was centered on the lens cap. Subsequently, the horizontal angle was adjusted to center the image of the central pinholes on the sensor. After a few iterations of the previous two steps the camera was aligned.

Ideally, when changing the lens, only the tripod needs to be translated along the laser line in order to adapt to the new focal lens. In practice, the horizontal angle and position need to checked and slightly realigned when the tripod is shifted.

## 2.3. Data collection and PSF extraction

Fully automated data acquisition was performed using a bash-script running a series of `gphoto2`[3] commands. Prior to the actual data collection the lens was focused as follows: A series of images across the approximate focus position was taken. For each image the RMSE of the 16 most central peaks in the green colour channel was calculated. The focus position with the minimal RMSE was then used for subsequent data acquisition.

The data set consists of four aperture settings for each lens: open aperture, 2.8, 4, and 5.6. Especially in the case of open aperture the size of the PSF varies substantially when going from the centre to the edges of the image. Therefore, a series of different exposure times was taken for each aperture setting. Finally, in order to improve the signal to noise ratio, 10 images were taken for each lens and capture setting (i.e. aperture and exposure).

---

[3] http://www.gphoto.org

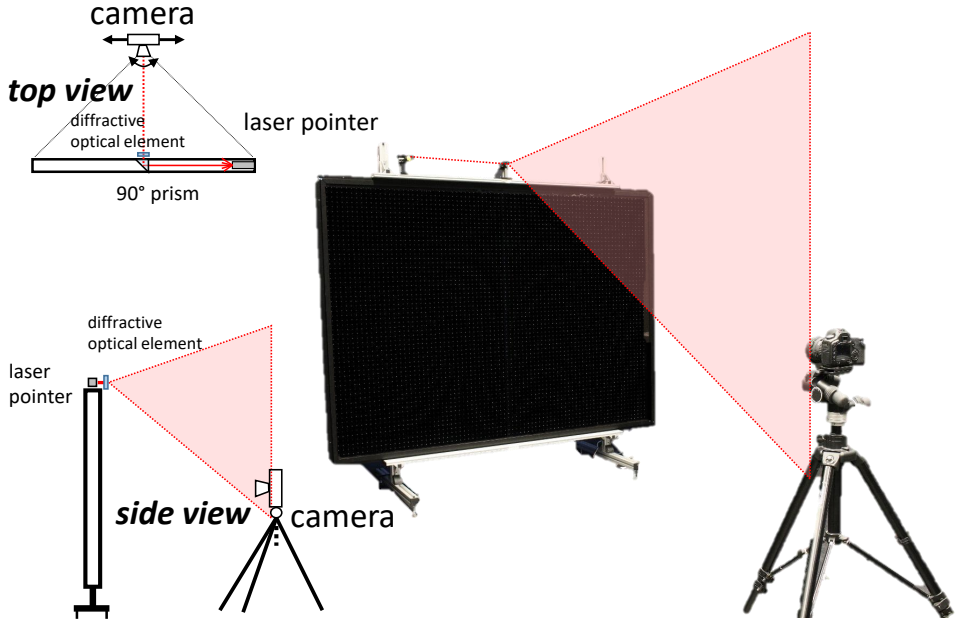

Figure 3. Alignment of the camera. The drawing of the top view illustrates the laser-line assembly. The laser-line projected from the center of the upper edge of the panel helps to horizontally center the camera.

All images of the PSF-panel were taken in the camera raw format and subsequently developed into tiffs using `dcraw`[4]. For details of the RAW development, see Sec. 3. The developed images were averaged over the 10 images per setting and the positions of the PSF-centroids were obtained using a procedure consisting of thresholding, eroding, and searching for connected regions. Out of all exposure times, the PSFs with the longest exposure times but without saturated pixel values within its $111 \times 111$ pixel patch were selected. The background of each patch was determined from its 4 corners. More precisely, the $5 \times 5$ pixel corner areas, that purely contained background even for large PSFs, were used to calculate the mean value $\mu$ and the standard deviation $\sigma$ of the background noise. The value of $\theta_{BG} = \mu + 4\sigma$ was then used as a threshold to segment the patch into PSF and background.

The PSFs were subsequently normalised such that the sum over all pixel values equals one.

---
[4] https://www.cybercom.net/~dcoffin/dcraw

### 2.4. Fast kernel regression for smoothing and interpolation

In order to further reduce noise and to interpolate PSFs at unobserved locations, we developed a sped-up version of the kernel regression algorithm by Hirsch and Schölkopf [1] that interpolates the PSF $h(\mathbf{x})$, $\mathbf{x} = (u, v; r, \varphi)$, at a new location $(r, \varphi)$ as

$$h(\mathbf{x}) = \frac{\sum_i h_i(\mathbf{x}_i) K(\mathbf{x} - \mathbf{x}_i)}{\sum_i K(\mathbf{x} - \mathbf{x}_i)} \quad (1)$$

where $K(\cdot)$ is a kernel function and $i$ runs over the individual pixels of the recorded PSFs. We assume that the kernel factorises over its dimensions and use a squared exponential kernel with lengthscale $\ell_x$ for each factor $x \in \mathcal{X} = \{r, \varphi, u, v\}$:

$$K(\mathbf{x}) = \prod_{x \in \mathcal{X}} K_x(x), \quad K_x(x) = \exp\left(-\frac{\|x\|^2}{2\ell_x^2}\right) \quad (2)$$

The index $i$ in Eq. (1) runs over all pixels of all recorded PSFs; to obtain the value of one pixel at local coordinates $(u, v)$ for one new PSF located in a patch at $(r, \varphi)$ the algorithm by Hirsch and Schölkopf [1] computes the covariance with every other recorded data point (every pixel of every recorded PSF). This is clearly infeasible for our data (4800 PSFs with $111 \times 111$ pixels $\approx 6 \cdot 10^8$ data points). However, by exploiting the product structure of the kernel, see Eq. (2), we can speed up this computation. First, we rotate each patch to a common coordinate system by locally rotating it by an angle $-\varphi$, similar to Fig. 1 but in place. The associated interpolation will destroy some of the sub-pixel information, however, we found the difference to be negligible. This approximation makes the individual PSFs independent of the global patch coordinate $(r, \varphi)$, that is, $h_i(\mathbf{x}_i) \approx h_i(u_i, v_i)$ (note that we still keep track of the index $i$), such that

$$h(\mathbf{x}) = \frac{\sum_i h_i(\mathbf{x}_i) K(\mathbf{x} - \mathbf{x}_i)}{\sum_i K(\mathbf{x} - \mathbf{x}_i)}$$
$$\approx \frac{\sum_i h_i(u_i, v_i) K_u(u_i - u) K_v(v_i - v)}{\sum_i K(\mathbf{x} - \mathbf{x}_i)} \times \quad (3)$$
$$\times K_r(r_i - r) K_\varphi(\varphi_i - \varphi)$$

Thus, we can compute $K_r(r_i - r) K_\varphi(\varphi_i - \varphi)$ for all PSF locations $(u, v)$ simultaneously. Moreover, as the local PSF coordinate $u, v$ are pixel indices (integers) and only a small number of combinations $|u - u_i| \in \{0, \ldots, 111\}$ exists, we can pre-compute and subsequently look up the contributions of $K_u(u_i - u) K_v(v_i - v)$. This pre-computation is only efficient because the squared exponential kernel is stationary (translationally invariant), that is, it only depends on the differences of the coordinates $u - u_i$ and $v - v_i$, not the individual values.

Naively, one might expect that this treatment of the local coordinates destroys all sub-pixel information. However, due to the interpolation when rotating $h_i$ to the common coordinate-frame, a large part of the sub-pixel information is actually maintained.

## 3. Data (pre-)processing and training data

In the following we present details about the data-(pre)processing pipeline.

### 3.1. Image capture

All photographs were taken with a Canon 5DS R camera body and different lenses as listed in Tab. 2. The images were recorded using a tripod and saved in the camera RAW format. For all shots the lowest ISO setting (ISO100) and mirror lock-up as well as a tripod were used to reduce noise and blur due to camera shake, respectively.

### 3.2. RAW development

The images were developed into linear 16bit TIFFs using the tool `dcraw` with flags `-T -4 -q 0 -o 0`. Thus, camera RAW colours were used and demosaicing/debayering was performed with bilinear interpolation. The latter ensures that colour channels are not mixed as lens aberrations vary with colour. To be consistent, we use the same settings for all RAW images, both for ground truth PSF images from the *PSF panel* and photographs of the regular pattern or natural scenes.

### 3.3. Patch extraction and rotation

For local MTF estimation with our method, we have to compute a forward pass of the image patches through the neural network. Before that, the image patches are rotated by an angle $-\varphi$, such that the radial and tangential direction align with the image axes. For this rotation, we extract suitably larger image patches that are rotated and subsequently cropped to the input size of the network. We used bilinear interpolation when rotating the images, as we found that more advanced interpolation methods can introduce unwanted artifacts (e.g. overshooting) that would be detrimental for MTF estimation.

### 3.4. Training data

In the following we give details about generation of the training data.

**Recorded ground truth lens PSFs.** The ground truth PSFs were collected as described in Sec. 2.3. As the PSFs are sampled on a regular grid, some radial locations occur more frequently than others. Especially the large PSFs in the corners of the image (large radii) are rare, in the sense that they only occur infrequently when randomly sampling PSFs from the grid. While the test set is similarly imbalanced, a single network trained with these PSF frequencies would specialise to predict particularly well on PSFs that it sees often during training and would generalise less well to rare PSFs. That is, the network would work well for small and round PSFs from the central regions of the image and less well on the corners. This setting is similar to training a discriminative learning system to perform classification with an imbalanced number of examples per class, which is a challenging problem in machine learning in general.

We circumvented this imbalance by using the fast kernel regression introduced in Sec. 2.4 to sample and interpolate PSFs such that all radii are represented equally, that is, we effectively reweighted the training data.

For training, we sampled PSFs from the following lenses and at the following settings:

- Zeiss Otus 55mm f/1.4 APO-Distagon @ f/1.4 and f/5.6

- Canon EF 24mm f/1.4L USM @ f/1.4 and f/2.8

- Canon EF 35mm f/1.4 USM @ 1.4

- Canon EF 50mm f/1.4 USM @ 2.8

- Canon EF 85mm f/1.2L II USM @ f/1.2, f/4, f/5.6

- Canon EF16-35mm f/2.8L II USM @ 16mm f/2.8 and f/4 (PSFs were recorded out of focus such that we did not use them to evaluate against this lens but we still used them as training data)

- Sigma 85mm f/1.4 (PSFs were recorded out of focus such that we did not use them to evaluate against this lens but we still used them as training data)

**Artificial PSFs.** In addition to real recorded PSFs, we also use artificial PSFs. These are composed of two Gaussians that model a central peak and a wider wing. The following parameters are chosen at random during training time: Size along the two main axes of the Gaussian for both Gaussians, relative weight of the two Gaussians.

**Sharp image patches: regular pattern.** The sharp image patches were created from a vector graphic representation of the regular pattern by converting it to raster graphics with different resolutions. Moreover, contrast and rotation were chosen at random.

**Sharp image patches: natural scenes** The sharp image patches were extracted from downscaled real photographs of natural scenes that have been captured with a sharp high-end lens, small aperture and under good light conditions using a tripod. We used the Zeiss Otus 55mm f/1.4 APO-Distagon lens and downscaled the images by a factor of two in each dimension to reduce the size of the effective PSF on these photographs. The photographs were developed into `tiffs` as described above, and subsequently synthetically blurred.

## 4. Details of the network

### 4.1. Subsampling into channels

We provide an example of how subsampling into channels works in Fig. 4. For more details, refer to Shi et al. [2].

### 4.2. Details of the network architecture

We present the exact architecture of our network in Tab. 1. The input is a single patch that has been rotated to the common coordinate system at $\varphi = 0$, see Fig. 1. At the beginning, we duplicate the image patch, rotate the second copy of the image by $-90°$, and pass both copies through the network independently, see the diagram in the main text for details. The architecture given here is for one of the two copies.

We used residual blocks as proposed by He et al. [3]. When the residual blocks change the feature or output dimension, a suitably strided $1 \times 1$ convolution with the matching number of output features was used for the residual connection. We used rectified linear (ReLU) activation functions except for in the last layer where we used a sigmoid to constrain the output to lie in $[0, 1]$. We did not use batch normalisation as preliminary experiments indicated reduced regression performance.

We used the Adam optimiser [4] and decay the learning rate in steps from $10^{-4}$ to $10^{-6}$.

We implemented the network in `tensorflow` [5] using `tensorpack` [6], a neural network training library built on top of `tensorflow`, that also provides code for the residual blocks.

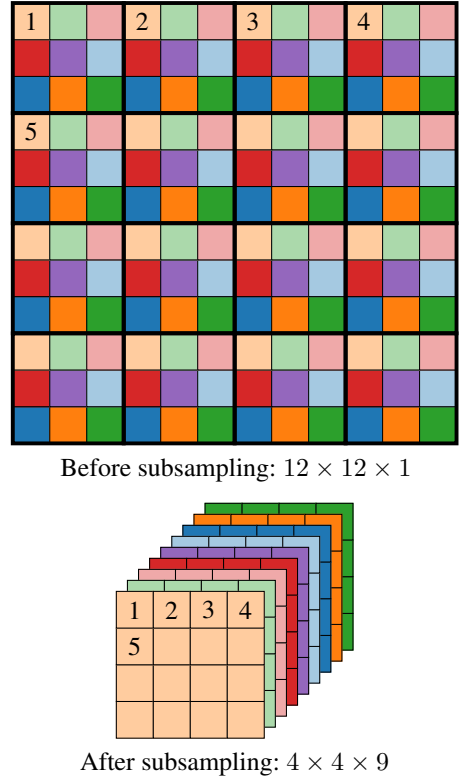

Before subsampling: $12 \times 12 \times 1$

After subsampling: $4 \times 4 \times 9$

Figure 4. Example of subsampling into channels. The input image consisting of individual pixels (coloured squares) is decomposed into non-overlapping groups of $3 \times 3$ pixels (thick squares). Every $n$th pixel in a group is moved to the $n$th channel in the subsampled image. Every output channel corresponds to a $3 \times 3$-fold downsampled version of the original image.

| Output size | Layers |
|---|---|
| $192 \times 192 \times 1$ | Input patch (green colour channel) |
| $192 \times 192 \times 2$ | Append gradient image |
| $32 \times 32 \times 72$ | Subsample using $6 \times 6$ grid |
| $32 \times 32 \times 128$ | Initial convolution $5 \times 5, 128$ |
| $32 \times 32 \times 128$ | Residual block $\begin{bmatrix} 5 \times 5, 128 \\ 5 \times 5, 128 \end{bmatrix}$ |
| $16 \times 16 \times 128$ | Residual block $\begin{bmatrix} 3 \times 3, 128 \\ 3 \times 3, 128 \end{bmatrix}$ |
| $8 \times 8 \times 256$ | Residual block $\begin{bmatrix} 3 \times 3, 256 \\ 3 \times 3, 256 \end{bmatrix}$ |
| $4 \times 4 \times 256$ | Residual block $\begin{bmatrix} 3 \times 3, 256 \\ 3 \times 3, 256 \end{bmatrix}$ |
| $2 \times 2 \times 256$ | Residual block $\begin{bmatrix} 3 \times 3, 256 \\ 3 \times 3, 256 \end{bmatrix}$ |
| $1 \times 1 \times 256$ | Residual block $\begin{bmatrix} 2 \times 2, 256 \\ 2 \times 2, 256 \end{bmatrix}$ |
| 256 | Intermediate feature representation |
| 256 | Fully connected layer 1 |
| 256 | Fully connected layer 2 |
| 128 | Fully connected layer 3 |
| 8 | Output |

Table 1. Details of our network architecture.

## 5. Extended analysis

In this section we present preliminary experiments, which explore the robustness of our method to noise as well as to the orientation of edges.

### 5.1. Robustness to noise

We briefly explored robustness to noise on synthetically blurred patches of the regular pattern. For this, we used the network to predict the MTF values of patches that have been blurred and, in addition, been corrupted with noise of varying size. We synthetically blurred the same patch of the regular pattern with a series of artificial blurs (sum of two Gaussians) of varying sizes. The blurred images were then pixelwise corrupted with iid Gaussian noise of varying amplitude. For each series of blurs, that is for all images with the same noise amplitude, we computed the average error between ground truth and predicted MTF values. Fig. 5 shows this error against the noise amplitude. Our method is robust to small to medium amounts of noise but fails for very large noise. We note that the data range is $[0.2, 0.8]$ such that a Gaussian noise with $\sigma = 0.05$ is already quite large and leads to very visible corruption.

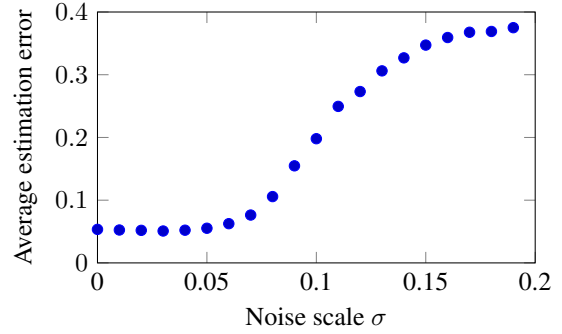

Figure 5. Average estimation error on a series of synthetically blurred patches of the natural pattern vs standard deviation of iid Gaussian pixel noise that was used to corrupt the blurred patches.

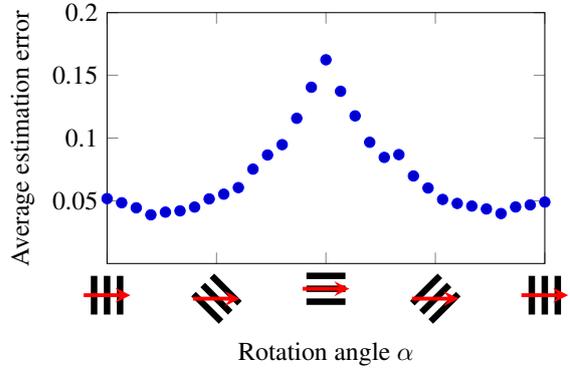

Figure 6. Average estimation error on a series of synthetically blurred patches of a stripe pattern at different angles. MTF estimation was performed in the horizontal direction (⟶) whereas the pattern is rotated from $0°$ to $180°$.

### 5.2. Orientation of edges

We also briefly explored the dependency on the orientation of edges with respect to the direction in which the MTF is estimated. As the regular pattern by Joshi et al. [7] is constructed to contain edges in all directions, we used a simple black and white stripe pattern instead. While the network has only been trained on the regular pattern, we found that it also worked well on stripes. In the experiment, we estimated the MTF in the horizontal direction but rotated the stripe pattern to different angles before it was blurred. Thus, in some cases the stripes were perpendicular to the direction of MTF estimation and in some cases they were parallel to it, see Fig. 6 for the results. The estimation error was smaller when the edges were perpendicular to the direction of MTF estimation and increased when both directions were co-aligned. This observation highlights the importance of texture with edges in several directions for reliable MTF estimation.

# 6. Example motives used for MTF estimation

In Fig. 7 we show examples of photos of natural scenes used to estimate the MTF from. The motives are selected to be planar and have structure (edges and sharp texture) across the entire field of view. In Fig. 8 we show example photographs of the regular pattern proposed by Joshi et al. [7].

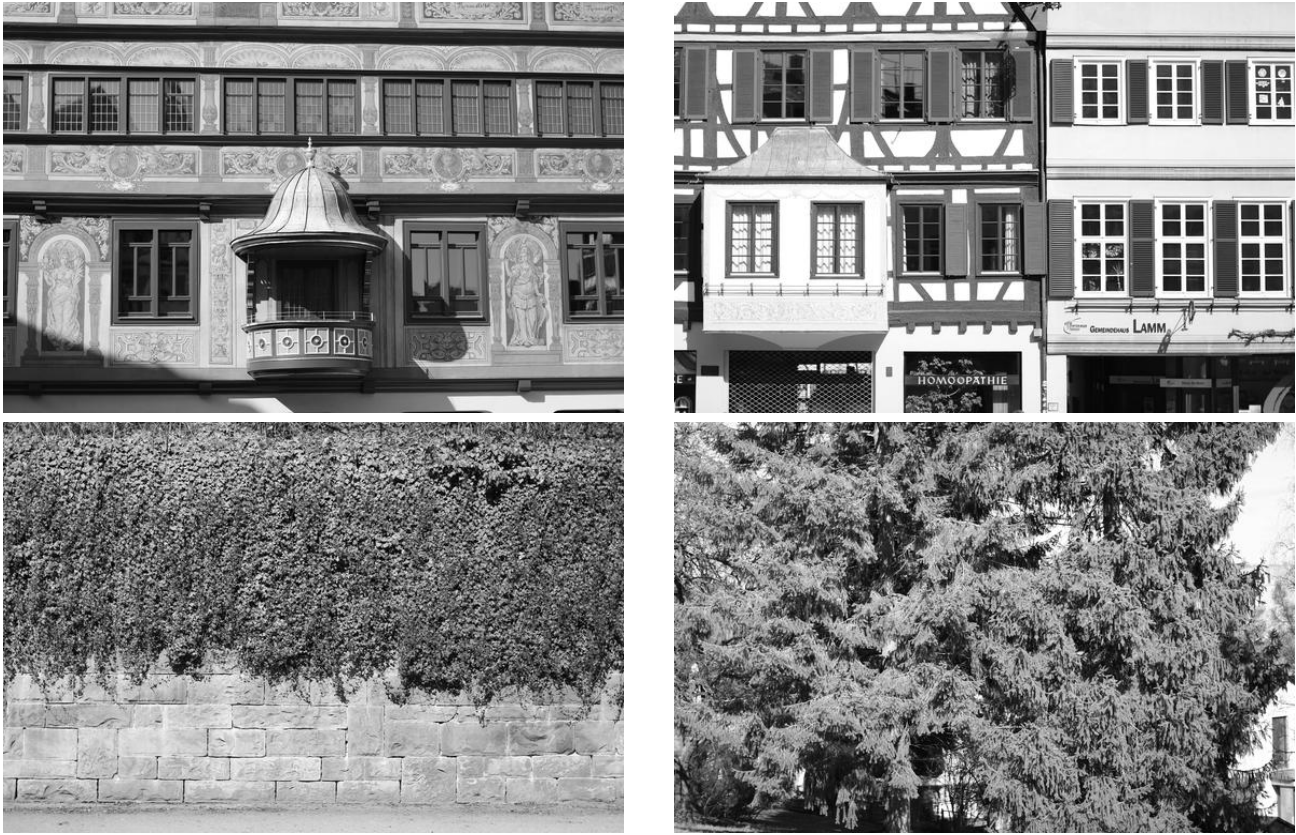

Figure 7. Examples of photographs of natural scenes used to estimate the MTF from.

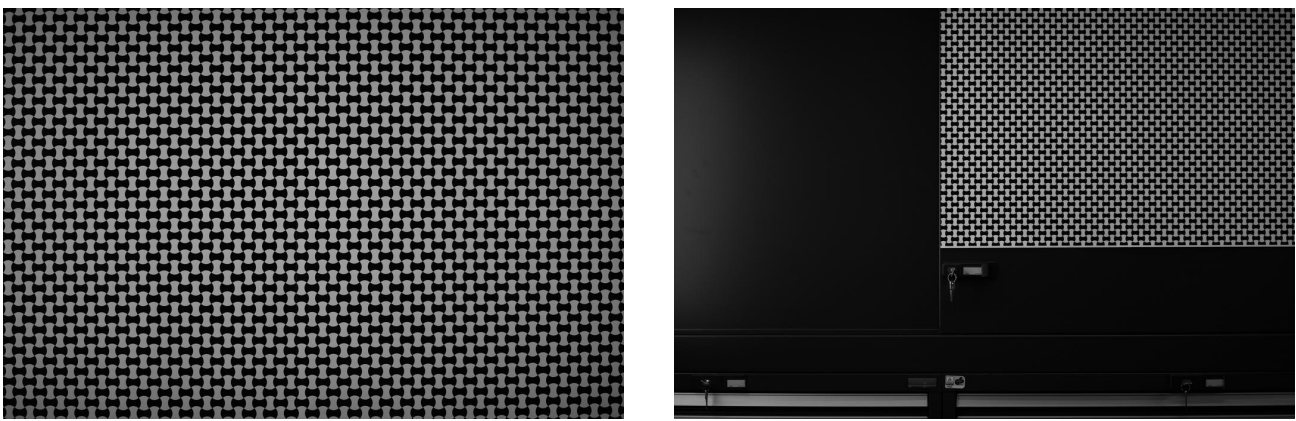

Figure 8. Examples of photographs from the regular pattern used to estimate the MTF from.

# 7. Extended measurements and results

In this section we present extended ground truth measurements of PSFs and corresponding MTFs as well as MTF estimates with our network for a series of consumer lenses. Tab. 2 gives an overview of the results. The graphs have the same structure as in the main text. First, we present extended results on ground truth measurements of PSFs and MTFs for most lenses. Then, we present estimates from photographs of the regular pattern or natural scenes sorted by lens.

Unless otherwise stated ("artificially blurred imgs"), all estimates are from real photographs either of the regular pattern or of natural scenes.

| Lens | Ground truth PSFs and MTFs | Estimates from photographs |
| --- | :---: | :---: |
| Sigma 50mm f/1.4 EX DG HSM | Fig. 10 (page 22) | Figs. 13 and 14 (page 25 and 26) |
| Sigma 35mm f/1.4 DG HSM | Fig. 11 (page 23) | Fig. 15 (page 27) |
| Zeiss Otus 55mm f/1.4 APO-Distagon | Fig. 10 (page 22) | Figs. 16 and 17 (page 28 and 29) |
| Zeiss 100mm f/2 | Fig. 12 (page 24) | Fig. 18 (page 30) |
| Canon EF 24mm f/1.4L USM | Fig. 9 (page 21) | Figs. 19 and 20 (page 31 and 32) |
| Canon EF 28mm f/1.8 USM | Fig. 11 (page 23) | Fig. 21 (page 33) |
| Canon EF 35mm f/1.4 USM | Fig. 9 (page 21) | Figs. 22 and 23 (page 34 and 35) |
| Canon EF 50mm f/1.4 USM | Fig. 9 (page 21) | Fig. 24 (page 36) |
| Canon EF 85mm f/1.2L II USM | Fig. 10 (page 22) | Fig. 25 (page 37) |
| Canon EF 135mm f/2L USM | Fig. 11 (page 23) | Fig. 27 (page 39) |

Table 2. Overview of measurement results obtained with our PSF panel (ground truth) as well as results for blind MTF estimation from real photographs.

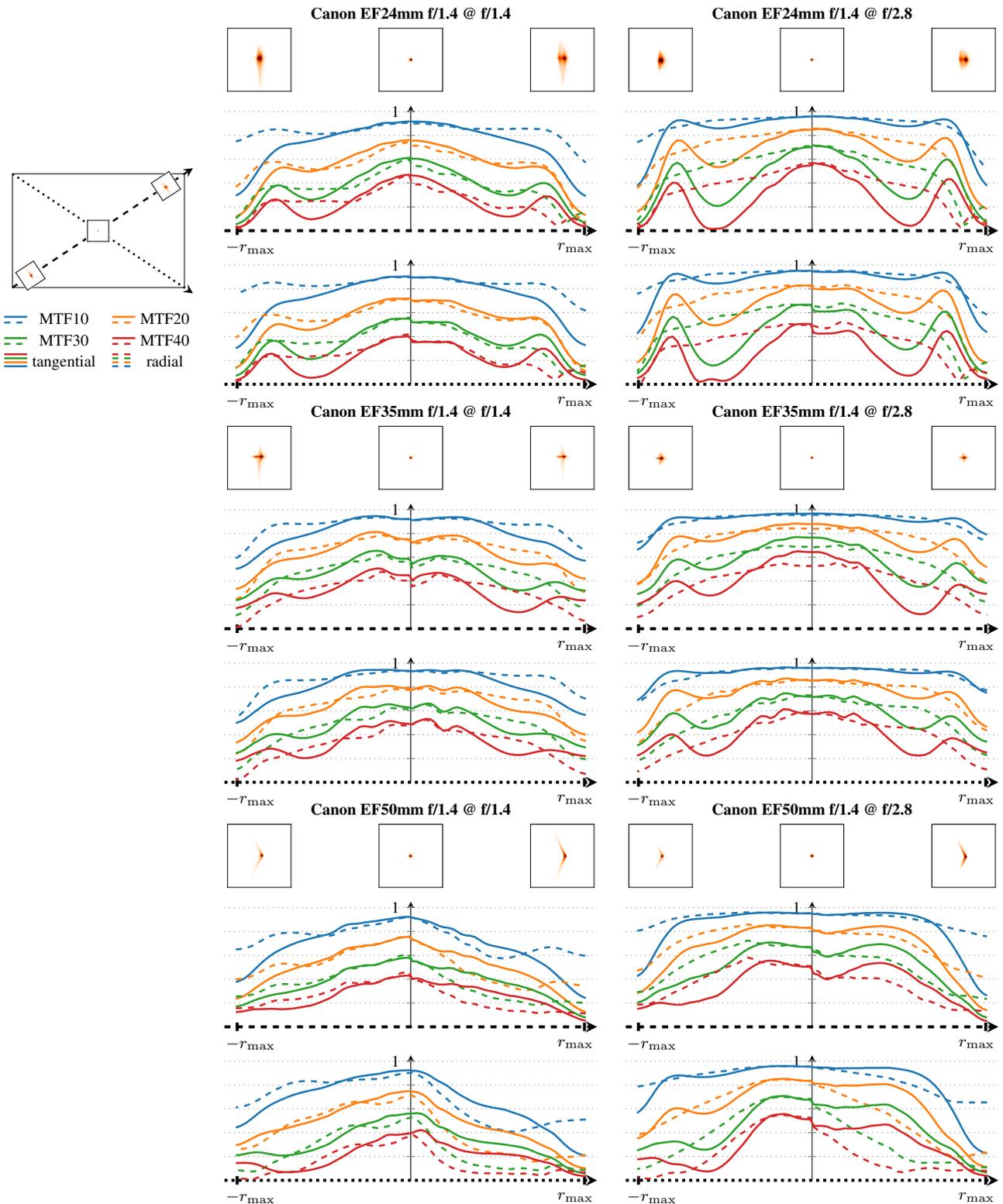

Figure 9. Ground truth global MTF charts in radial (⋯) and tangential (—) direction at $33.7°$ (top row, - - ▸) and $-33.7°$ (bottom row, ⋯▸). PSFs of size $111 \times 111$ pixels (top) are extracted at $-r_{\max}, 0, r_{\max}$ and rotated into the commen coordinate frame, see little cartoon at the top left.

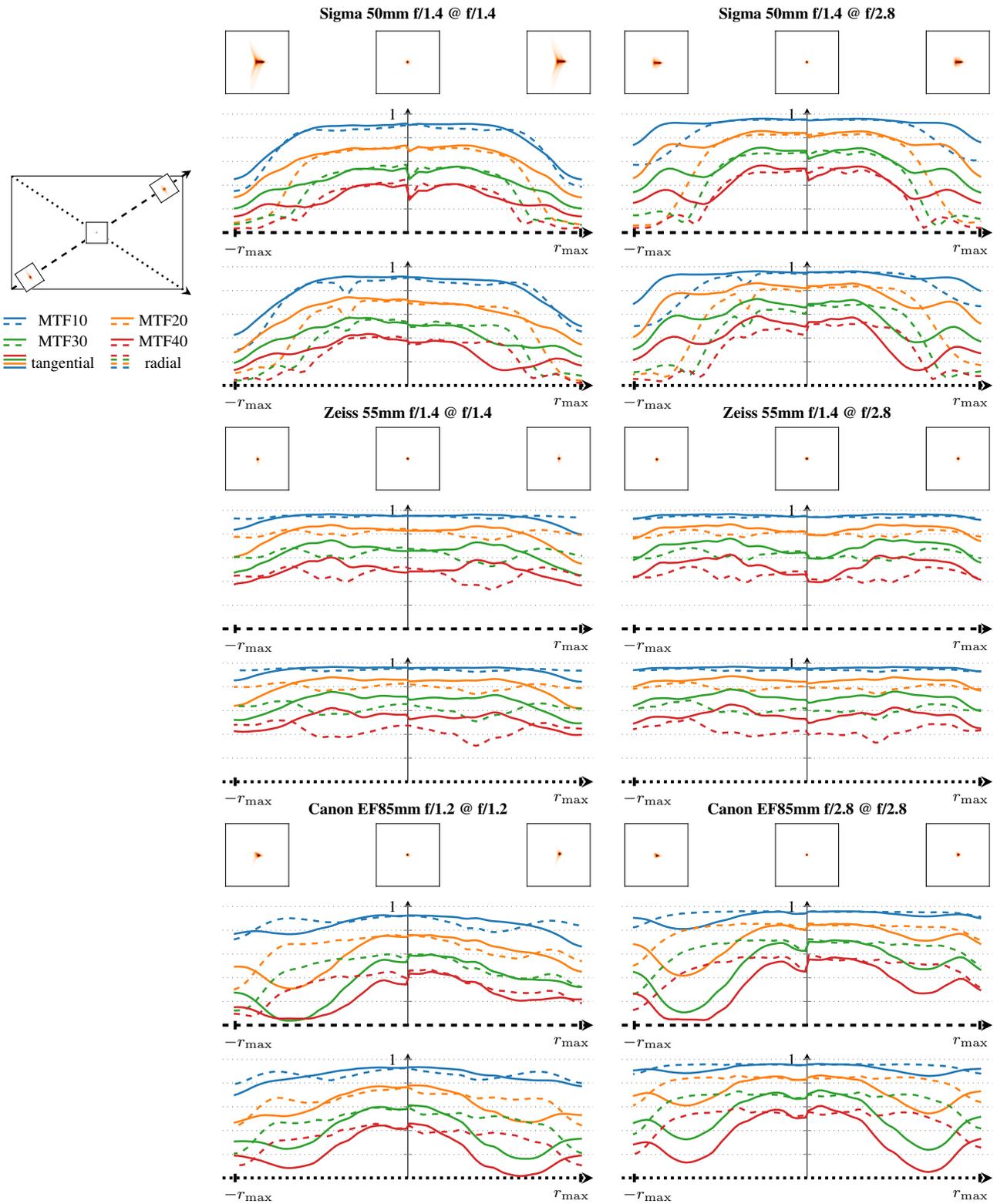

Figure 10. Ground truth global MTF charts in radial (┊┊┊) and tangential (≡) direction at $33.7°$ (top row, - -➤) and $-33.7°$ (bottom row, ⋯➤). PSFs of size $111 \times 111$ pixels (top) are extracted at $-r_{\max}, 0, r_{\max}$ and rotated into the commen coordinate frame, see little cartoon at the top left.

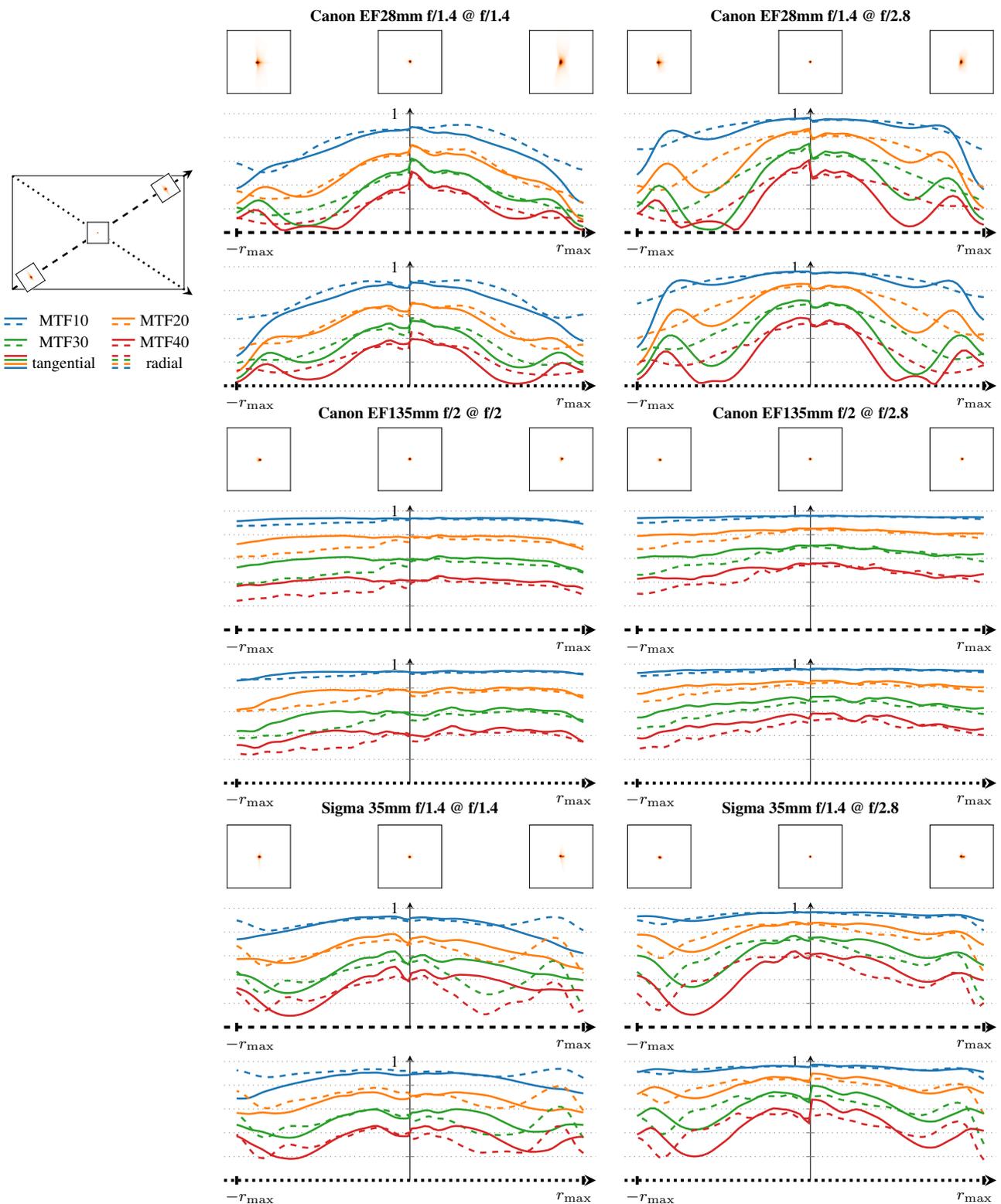

Figure 11. Ground truth global MTF charts in radial (┊┊┊) and tangential (━━) direction at $33.7°$ (top row, ╌╌➤) and $-33.7°$ (bottom row, ⋯➤). PSFs of size $111 \times 111$ pixels (top) are extracted at $-r_{\max}, 0, r_{\max}$ and rotated into the commen coordinate frame, see little cartoon at the top left.

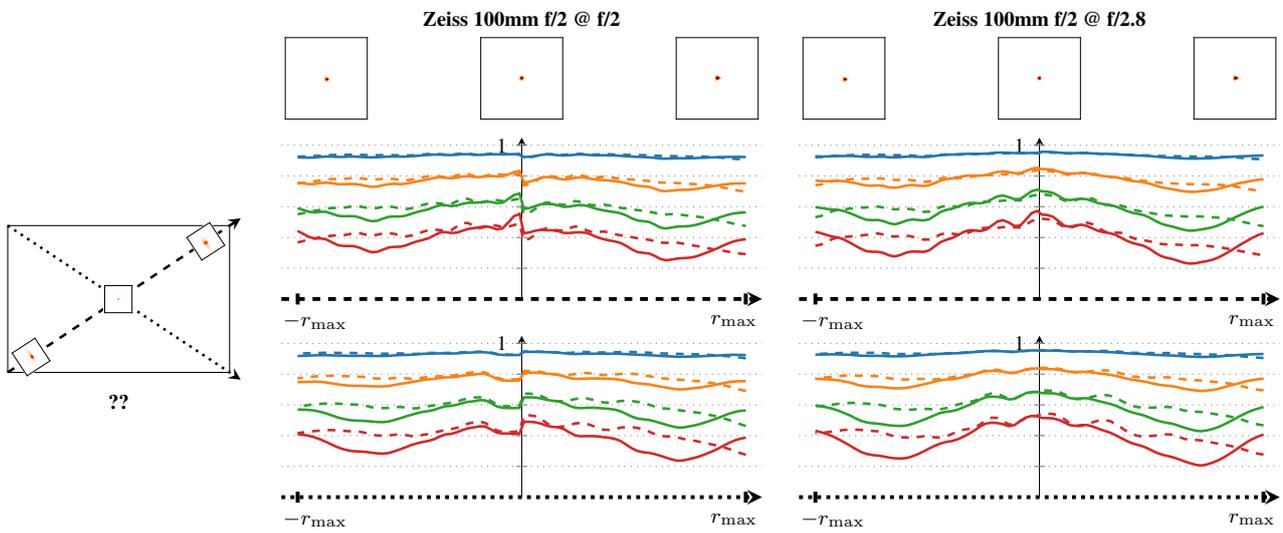

Figure 12. Ground truth global MTF charts in radial (┊┊┊) and tangential (═══) direction at $33.7°$ (top row, ╌╌➤) and $-33.7°$ (bottom row, ┈┈➤). PSFs of size $111 \times 111$ pixels (top) are extracted at $-r_{\max}, 0, r_{\max}$ and rotated into the commen coordinate frame, see little cartoon at the top left.

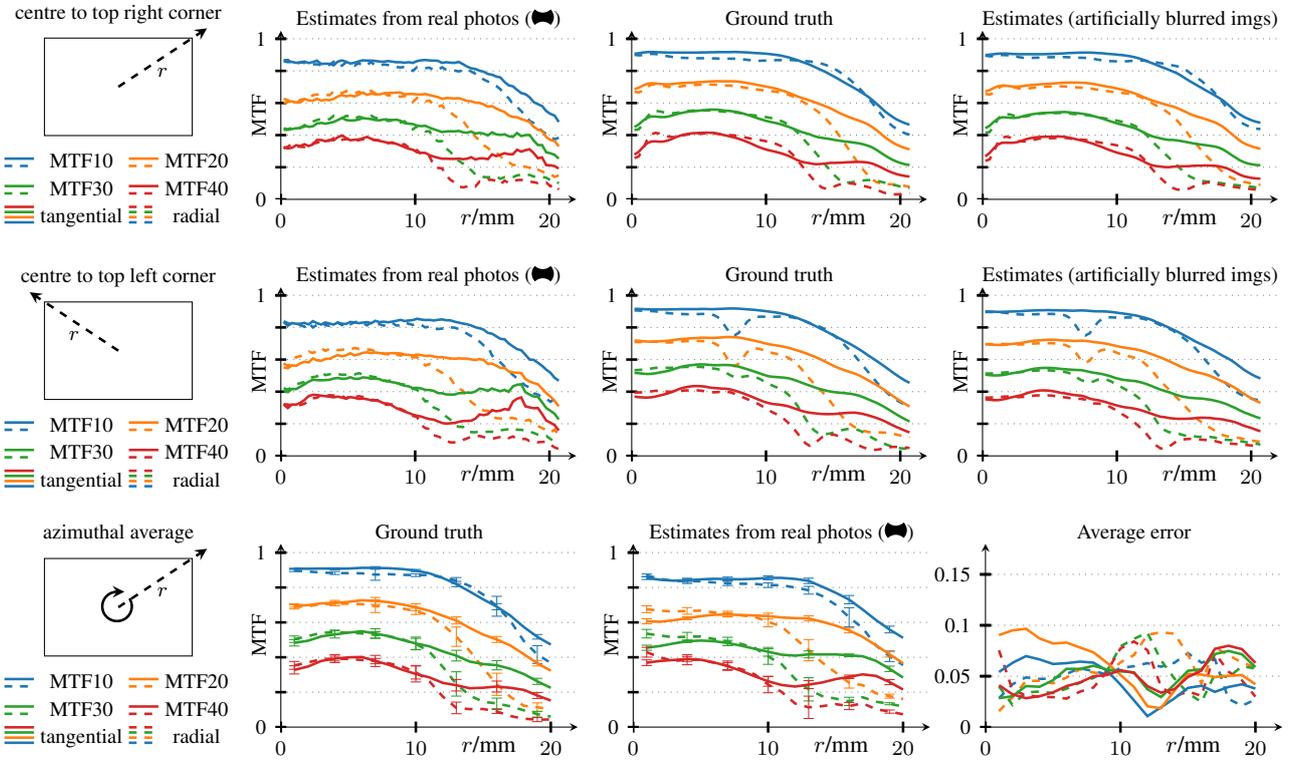
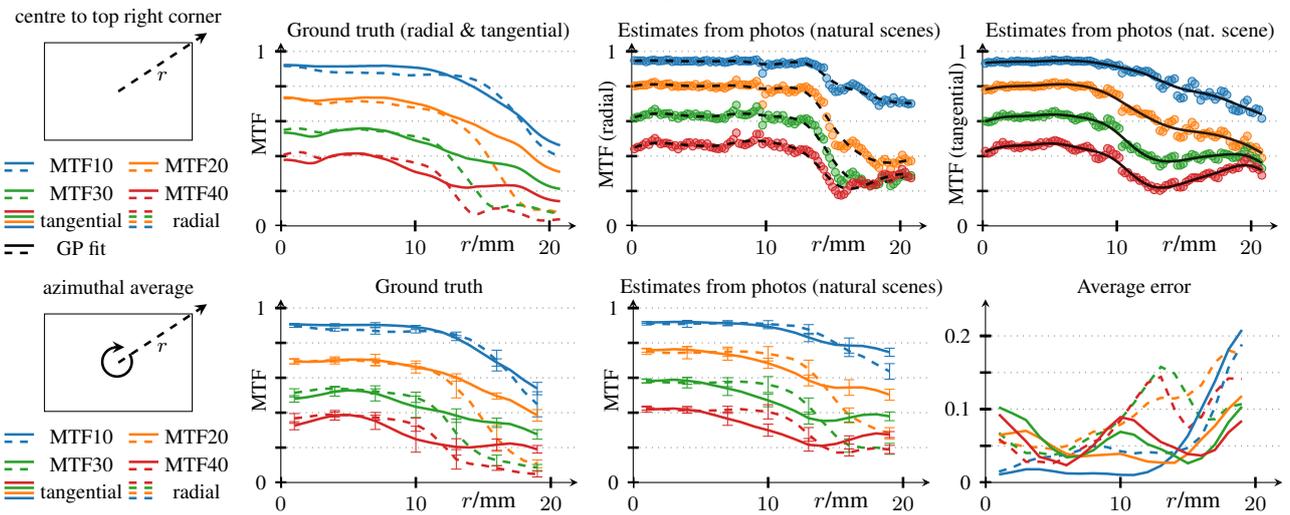

Figure 13. Results of MTF estimation for the regular pattern ⋈ (top three rows) and for natural scenes (bottom two rows) for the Sigma 50mm f/1.4 EX DG HS lens. Results are from the centre to the top left and top right corner as well as averaged over all angles (see little cartoon on the left).

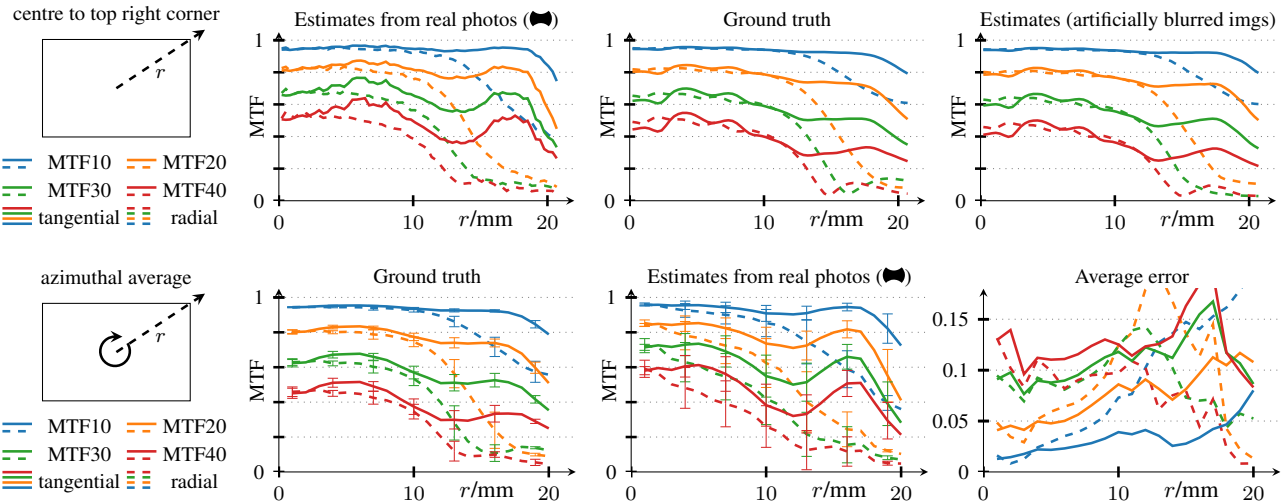
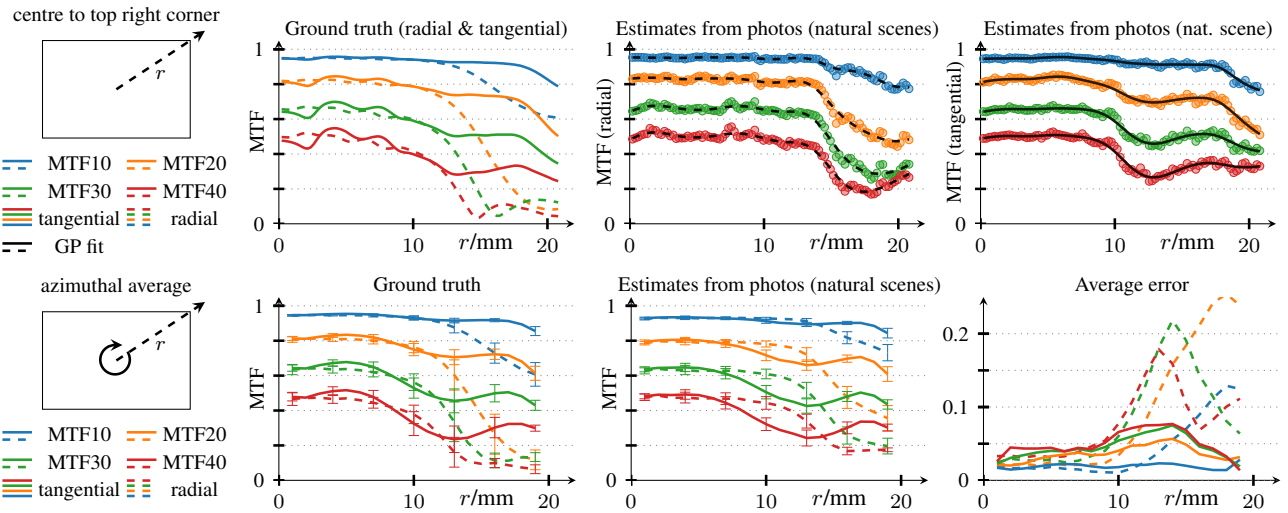
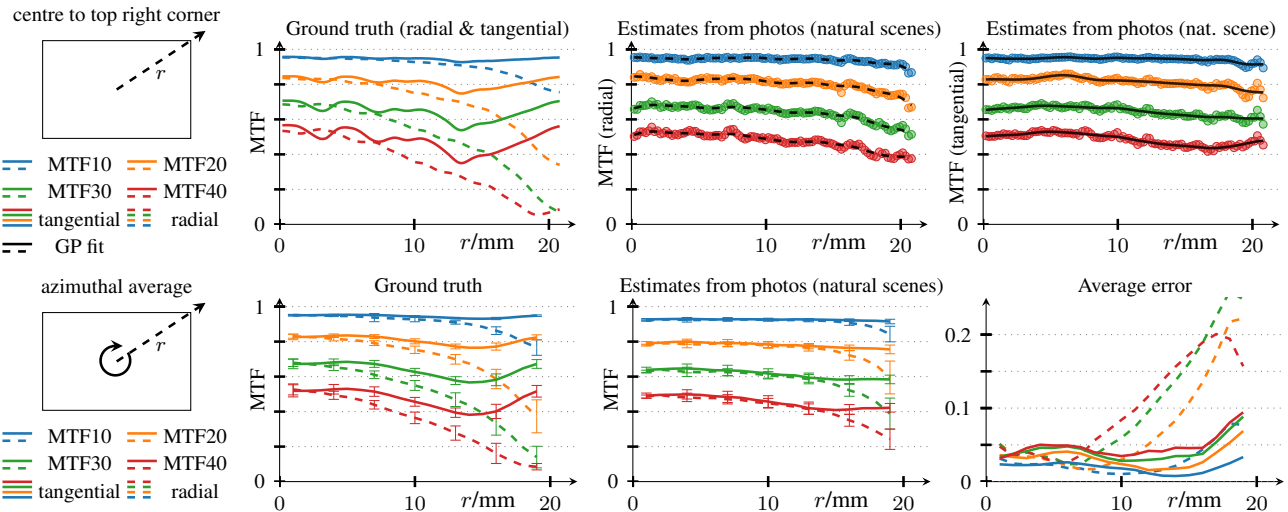

Figure 14. 28 Results of MTF estimation for the regular pattern ⋈ (top three rows) and for natural scenes (bottom two rows) for the Sigma 50mm f/1.4 EX DG HS lens. Results are from the centre to the top left and top right corner as well as averaged over all angles (see little cartoon on the left).

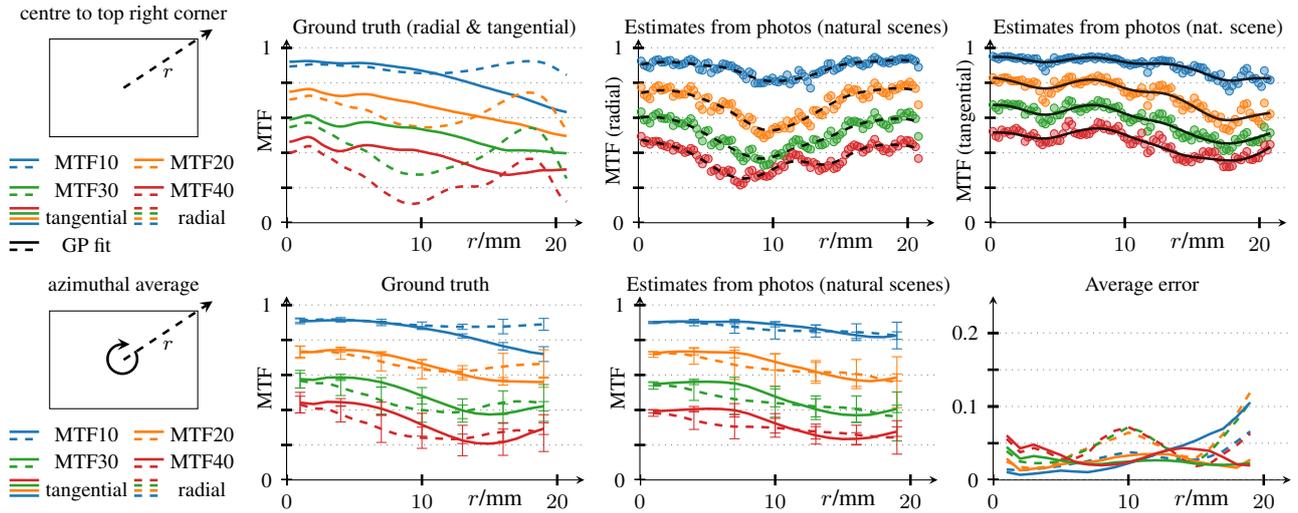
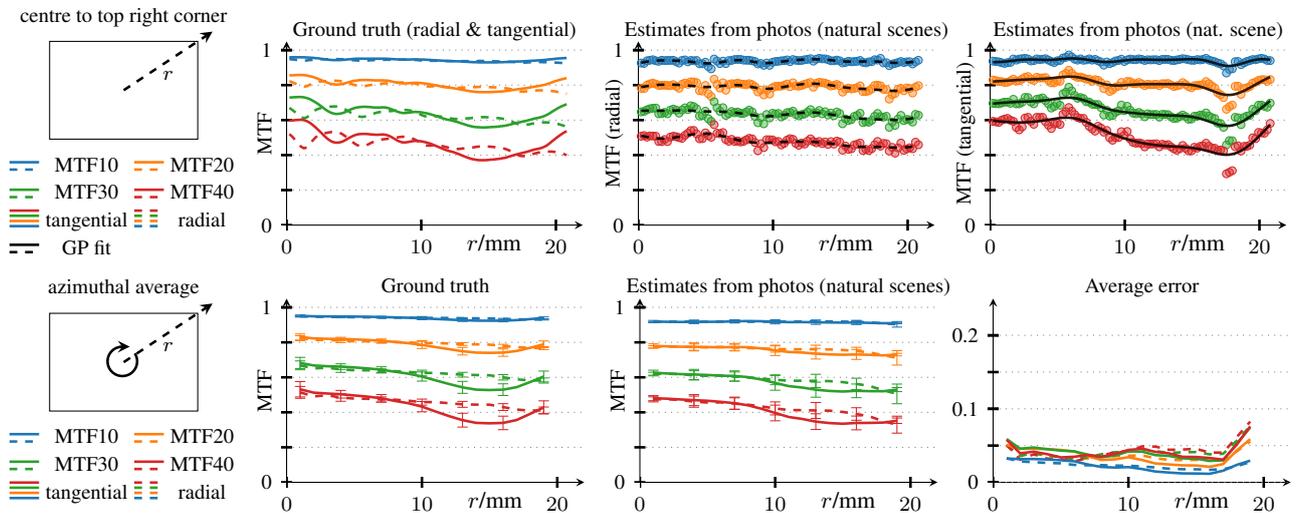
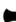

Figure 15. Results of MTF estimation for the regular pattern ⌘ (top three rows) and for natural scenes (bottom two rows) for the Sigma 35mm f/1.4 DG HSM lens. Results are from the centre to the top left and top right corner as well as averaged over all angles (see little cartoon on the left).

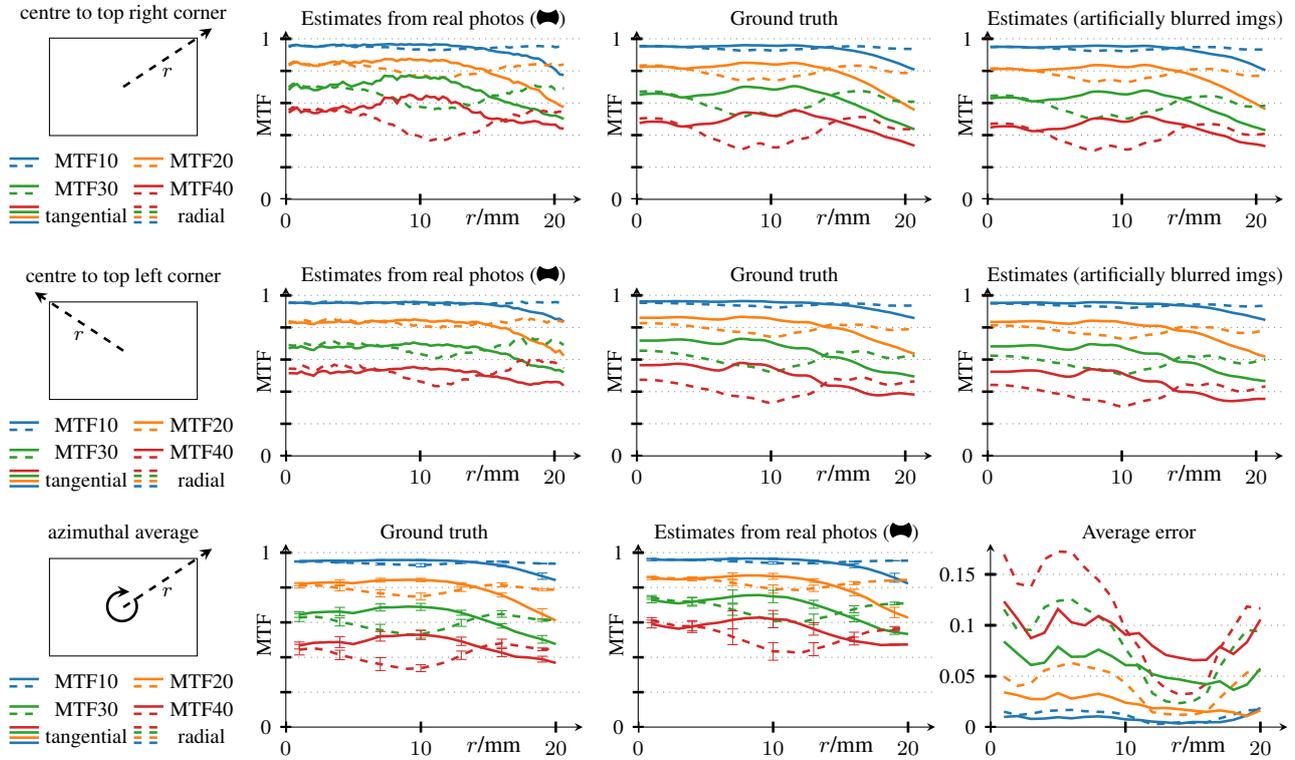
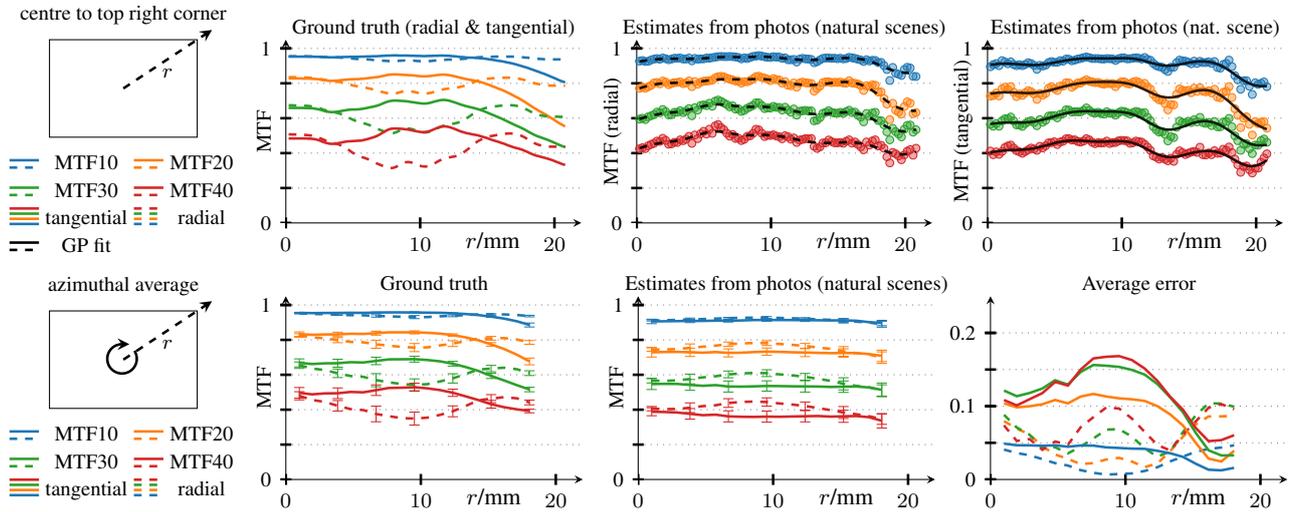

Figure 16. Results of MTF estimation for the regular pattern 🎀 (top three rows) and for natural scenes (bottom two rows) for the Zeiss Otus 55mm f/1.4 APO-Distagon. Results are from the centre to the top left and top right corner as well as averaged over all angles (see little cartoon on the left).

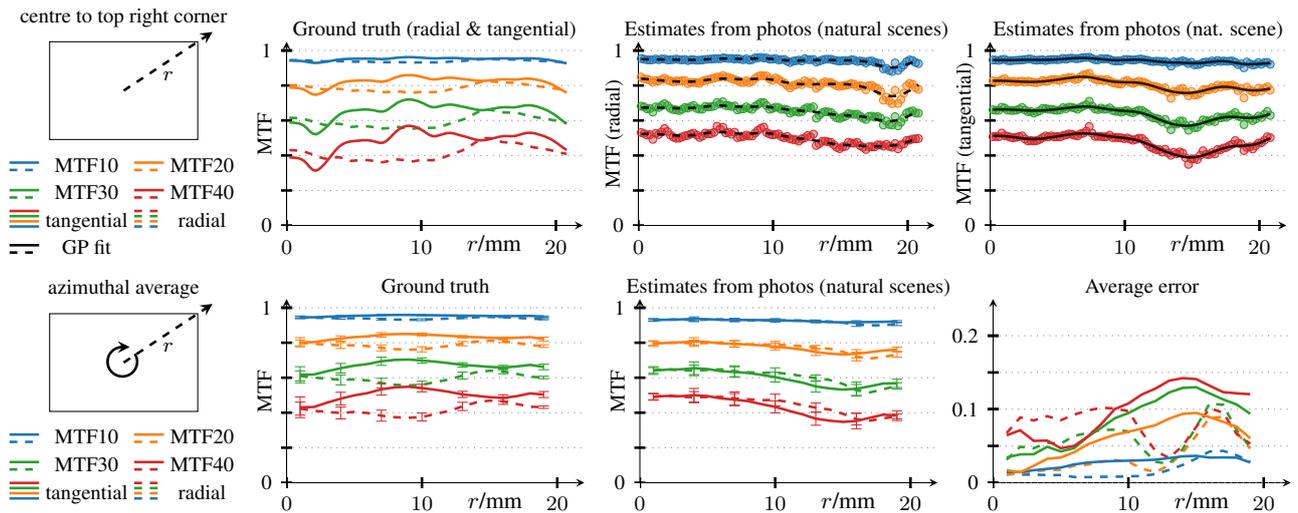
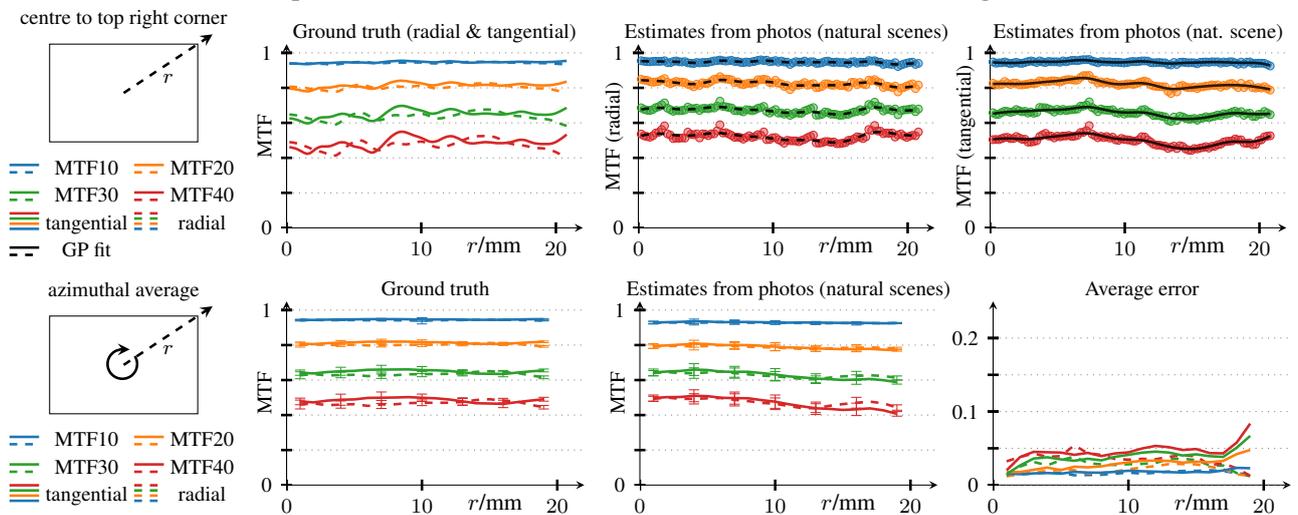

Figure 17. Results of MTF estimation for the natural scenes for the Zeiss Otus 55mm f/1.4 APO-Distagon. Results are from the centre to the top right corner as well as averaged over all angles (see little cartoon on the left).

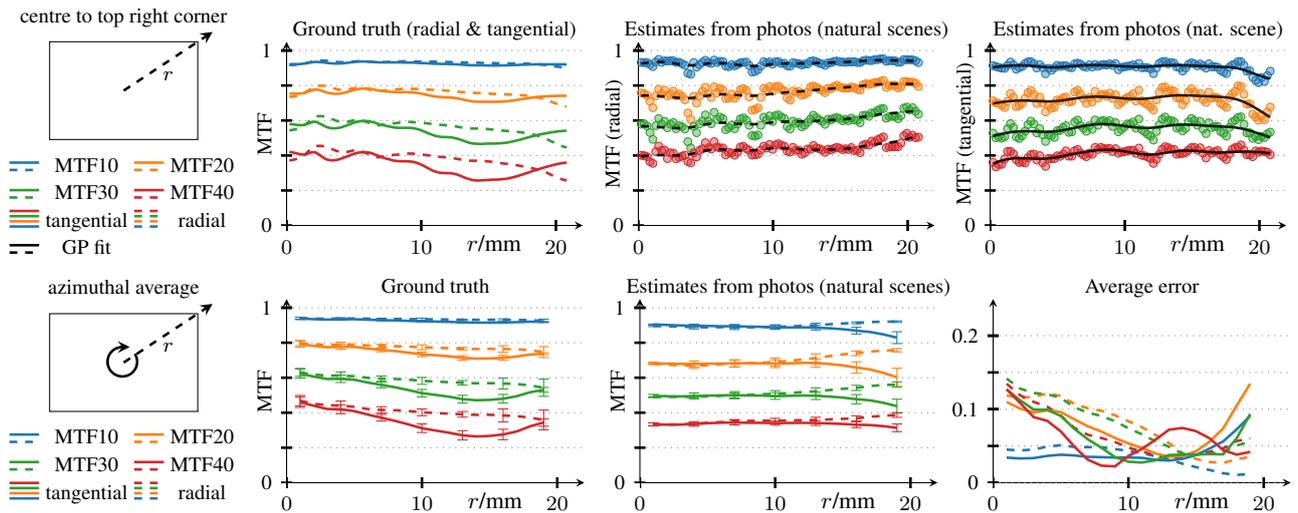

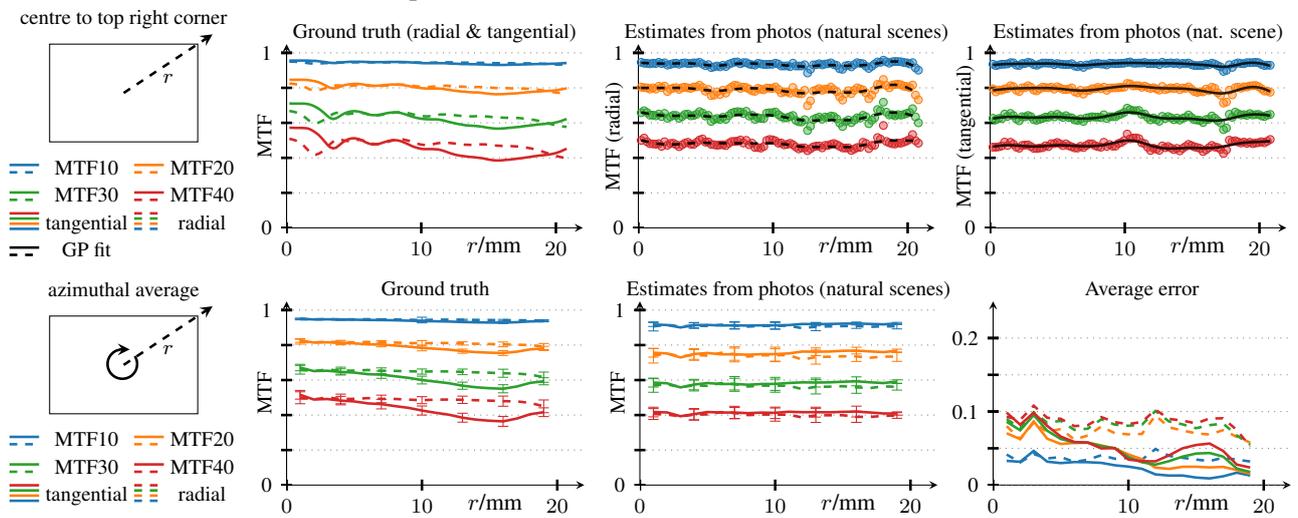

Figure 18. Results of MTF estimation for natural scenes for the Zeiss Otus 100mm f/2. Results are from the centre to the top right corner as well as averaged over all angles (see little cartoon on the left).

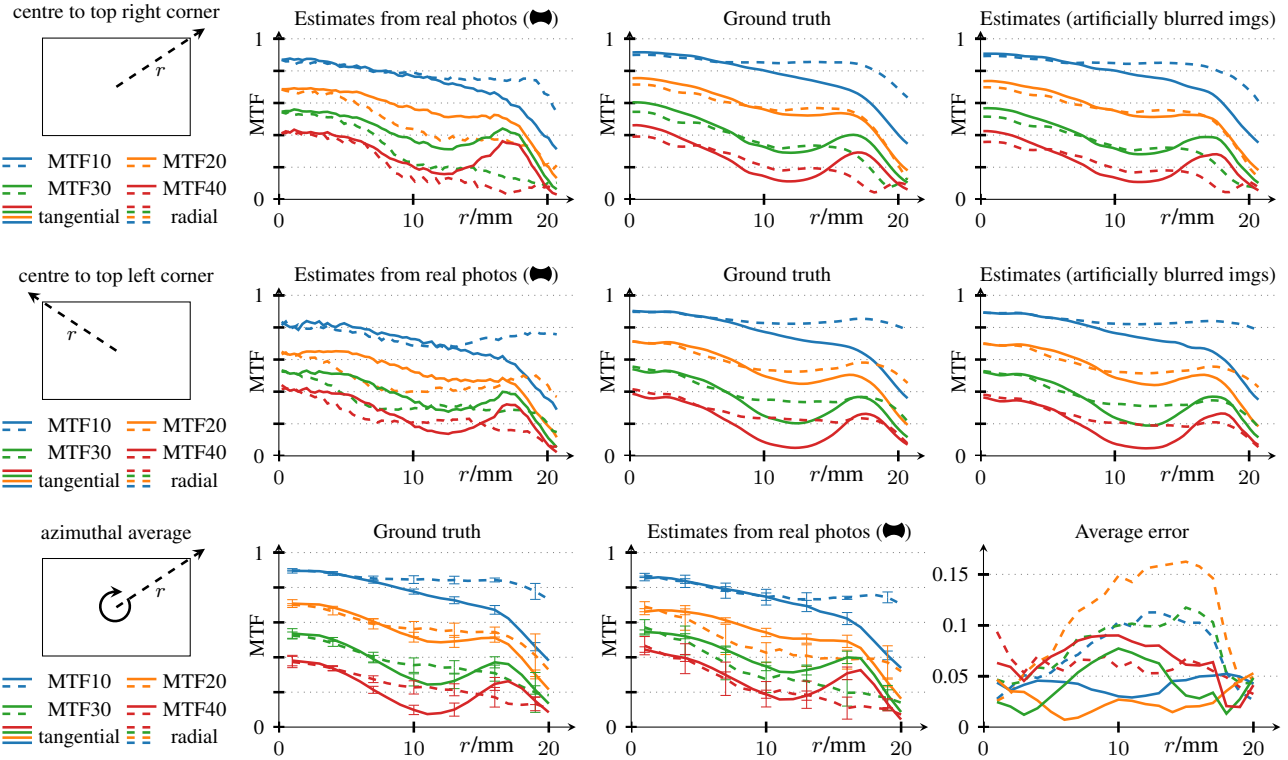
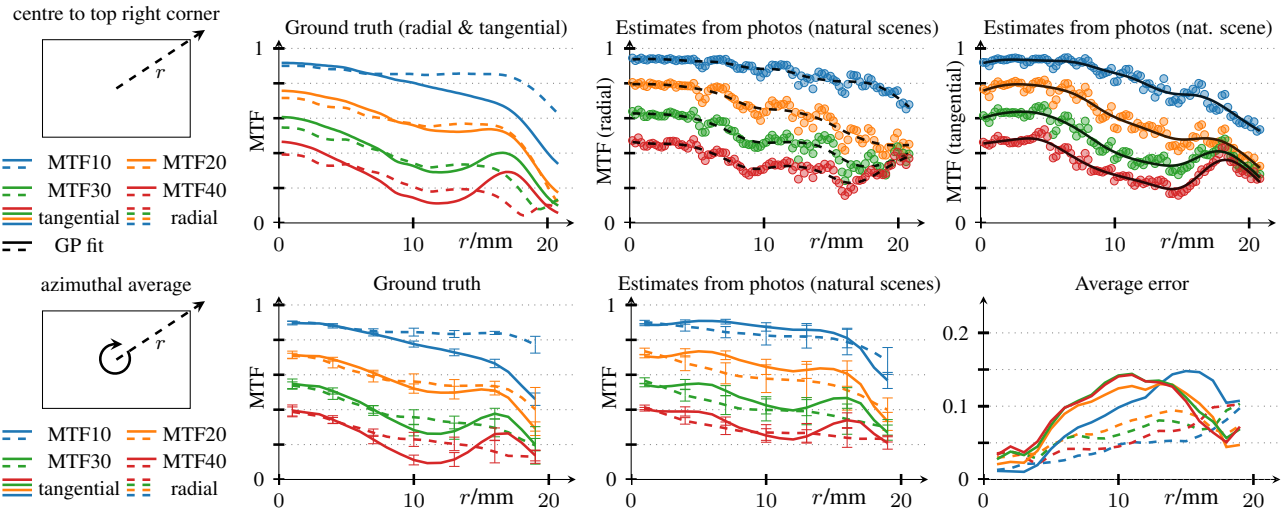

Figure 19. Results of MTF estimation for the regular pattern ⋈ (top three rows) and for natural scenes (bottom two rows) for the Canon EF 24mm f/1.4L USM. Results are from the centre to the top left and top right corner as well as averaged over all angles (see little cartoon on the left).

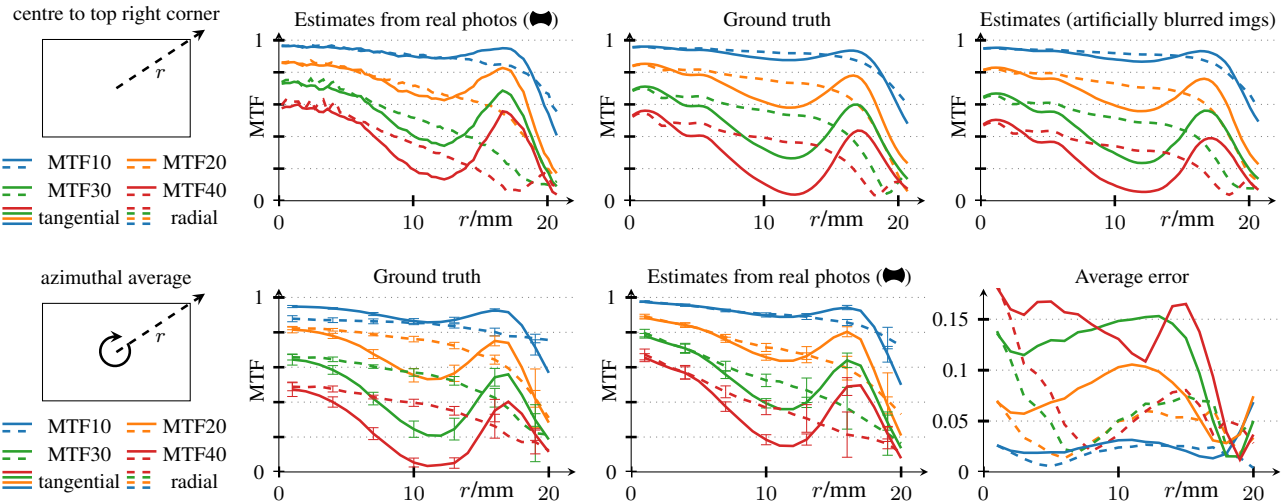
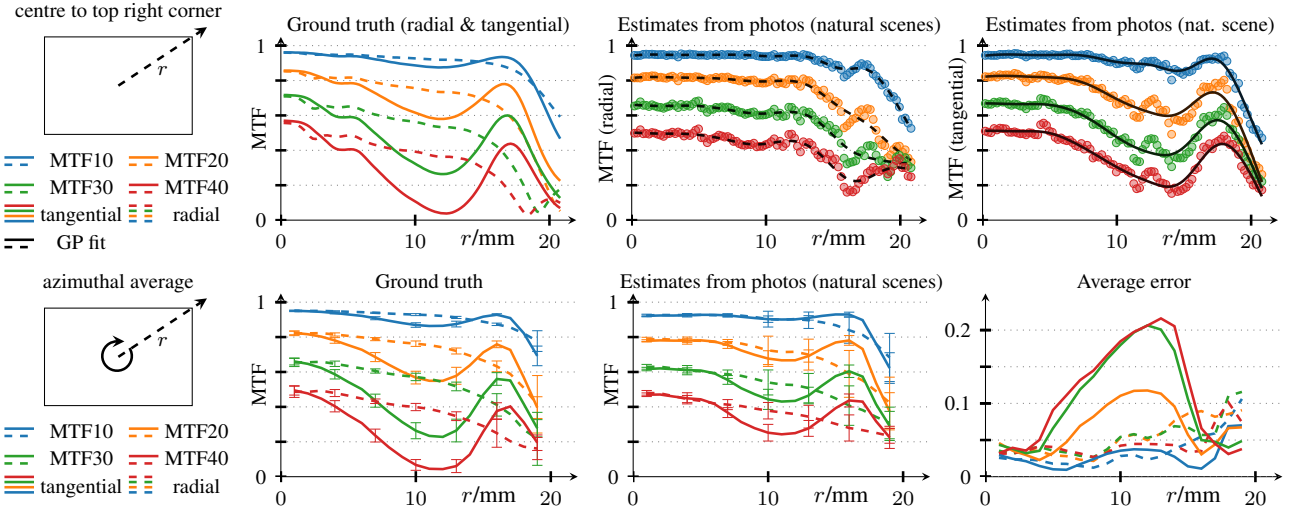
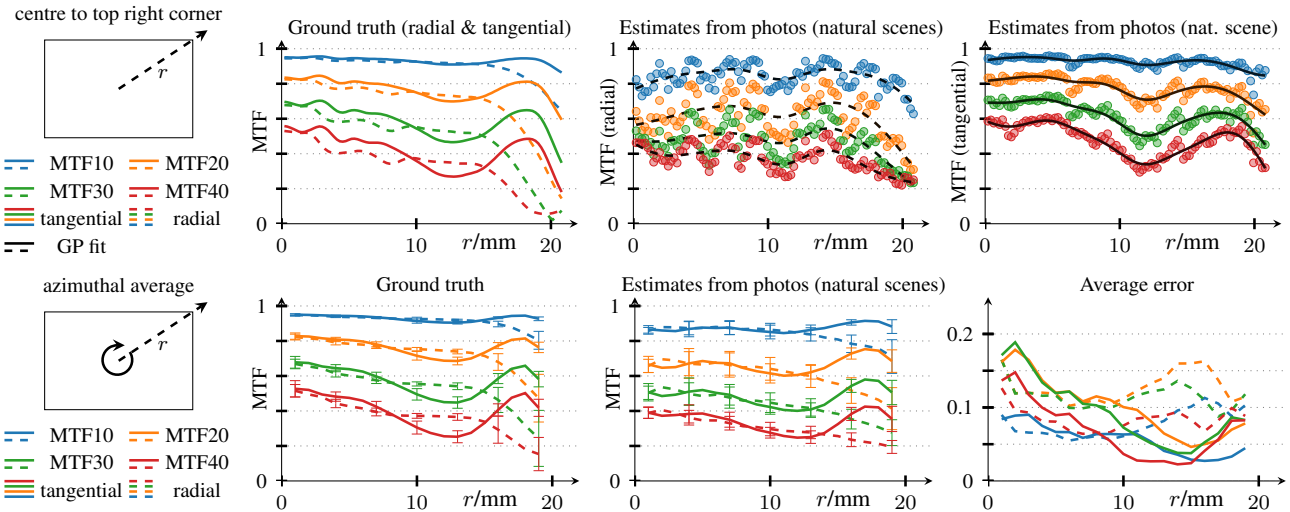

Figure 20. Results of MTF estimation for the regular pattern ⬛ (top three rows) and for natural scenes (bottom four rows) for the Canon EF 24mm f/1.4L USM. Results are from the centre to the top left and top right corner as well as averaged over all angles (see little cartoon on the left).

**Estimates from photos of natural scenes for Canon EF 28mm f/1.8 USM @ f/1.8**

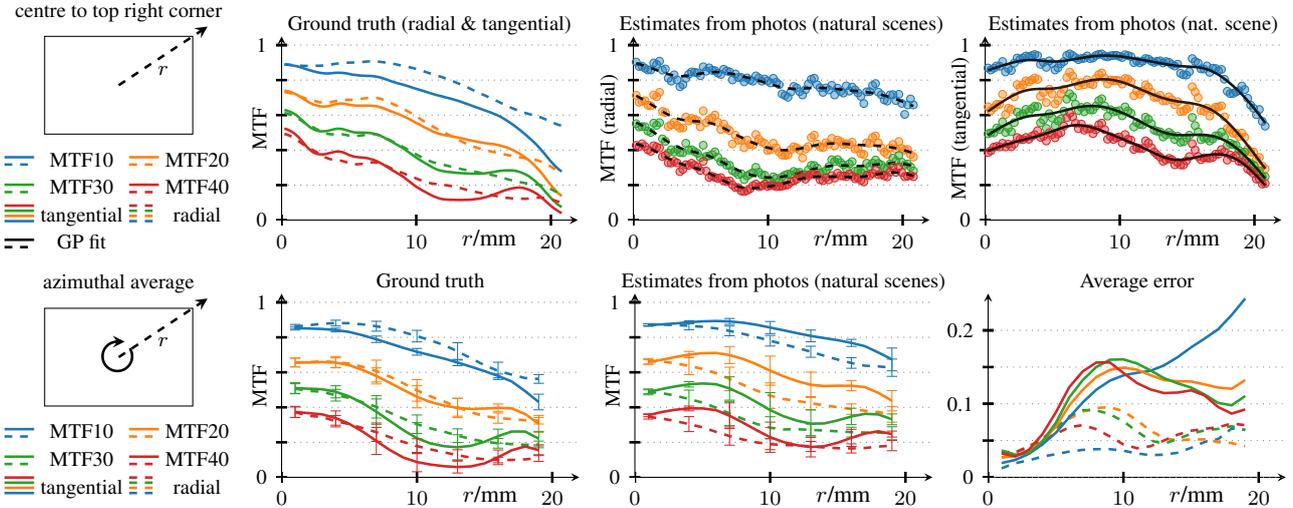

**Estimates from photos of natural scenes for Canon EF 28mm f/1.8 USM @ f/2.8**

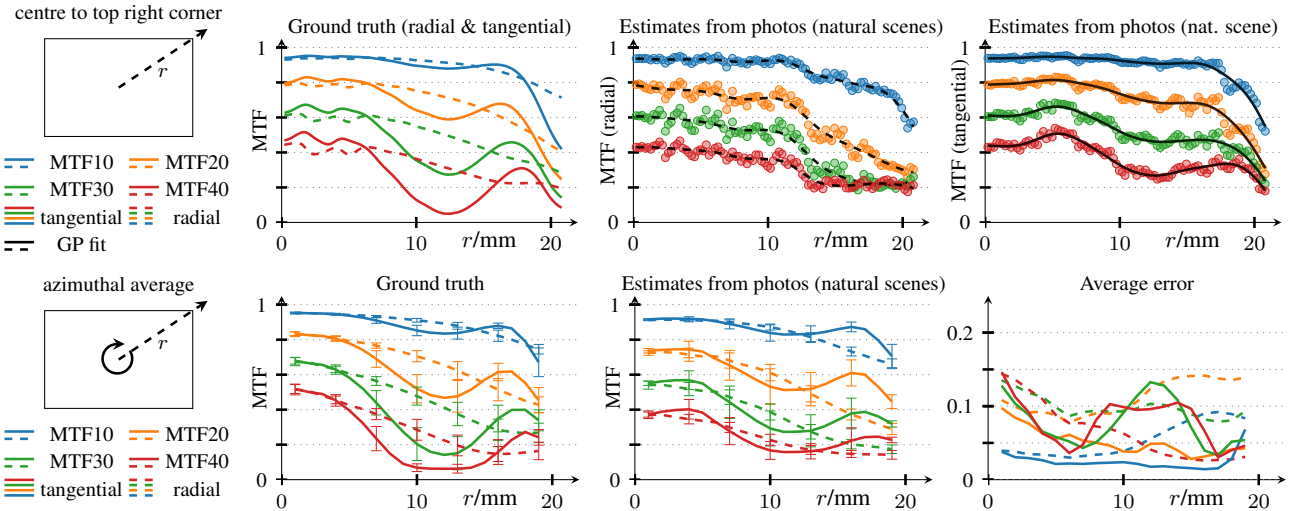

**Estimates from photos of natural scenes for Canon EF 28mm f/1.8 USM @ f/5.6**

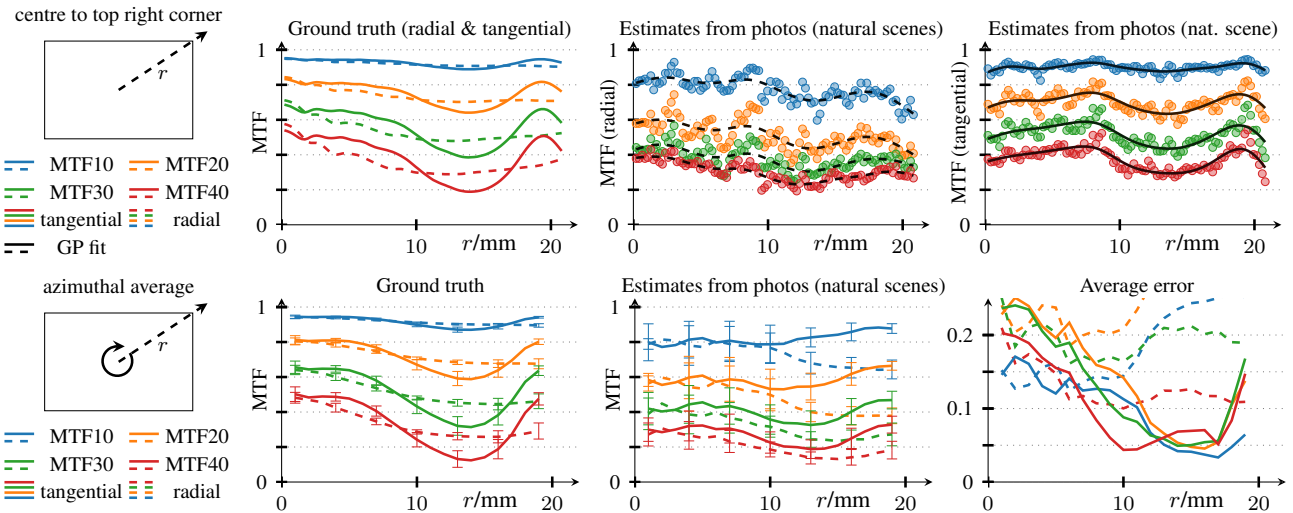

Figure 21. Results of MTF estimation for natural scenes for the Canon EF 28mm f/1.8L USM. Results are from the centre to the top right corner as well as averaged over all angles (see little cartoon on the left).

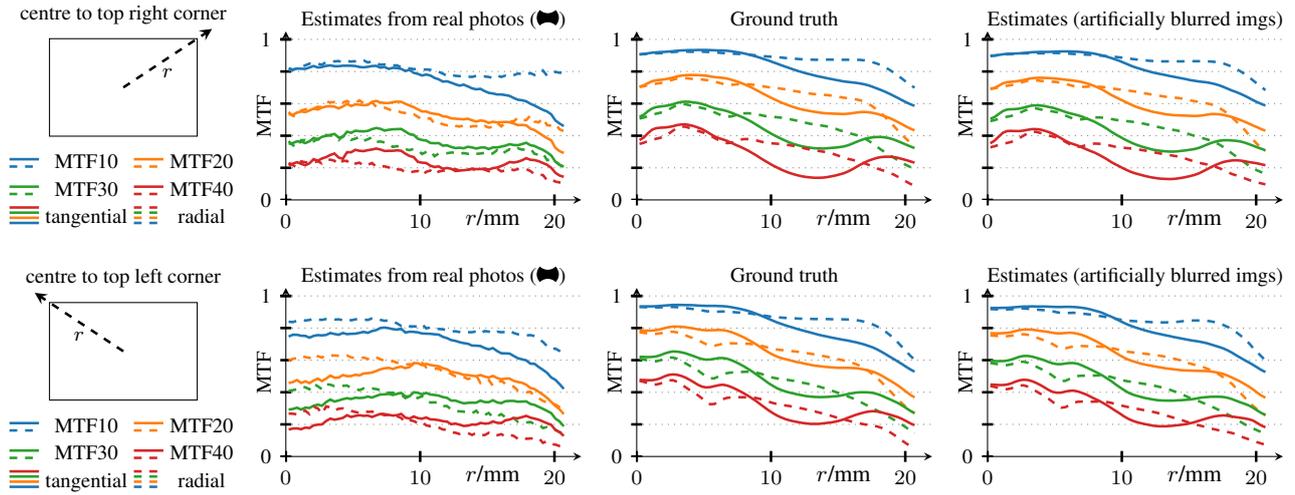
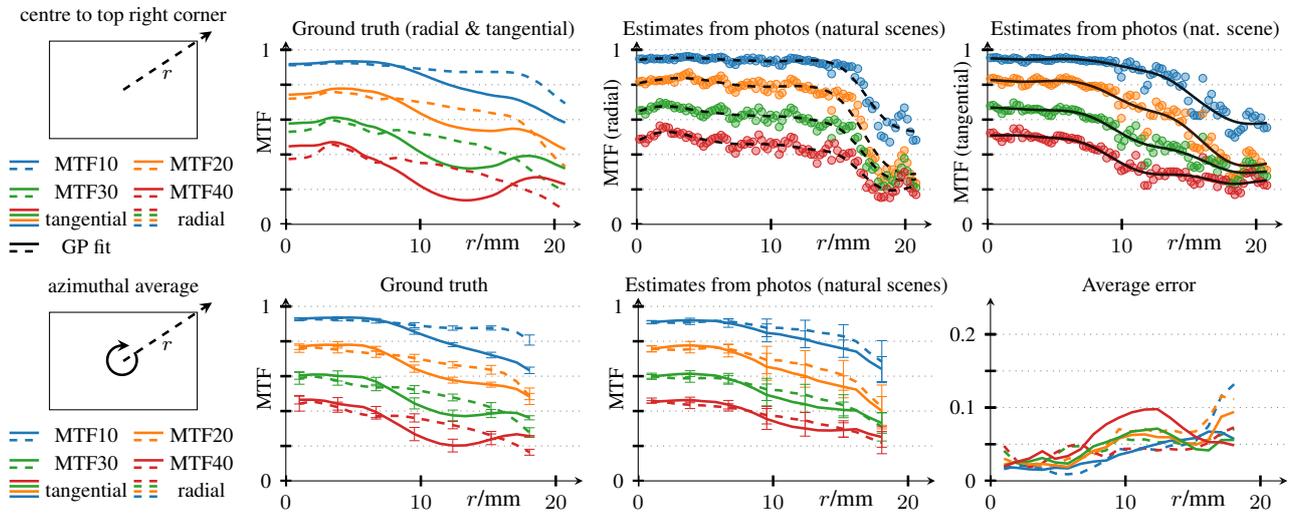

Figure 22. Results of MTF estimation for the regular pattern 🎀 (top three rows) and for natural scenes (bottom two rows) for the Canon EF 35mm f/1.4 USM lens. Results are from the centre to the top left and top right corner as well as averaged over all angles (see little cartoon on the left).

**Estimates from photos of the regular pattern for Canon EF 35mm f/1.4 USM @ f/2.8**

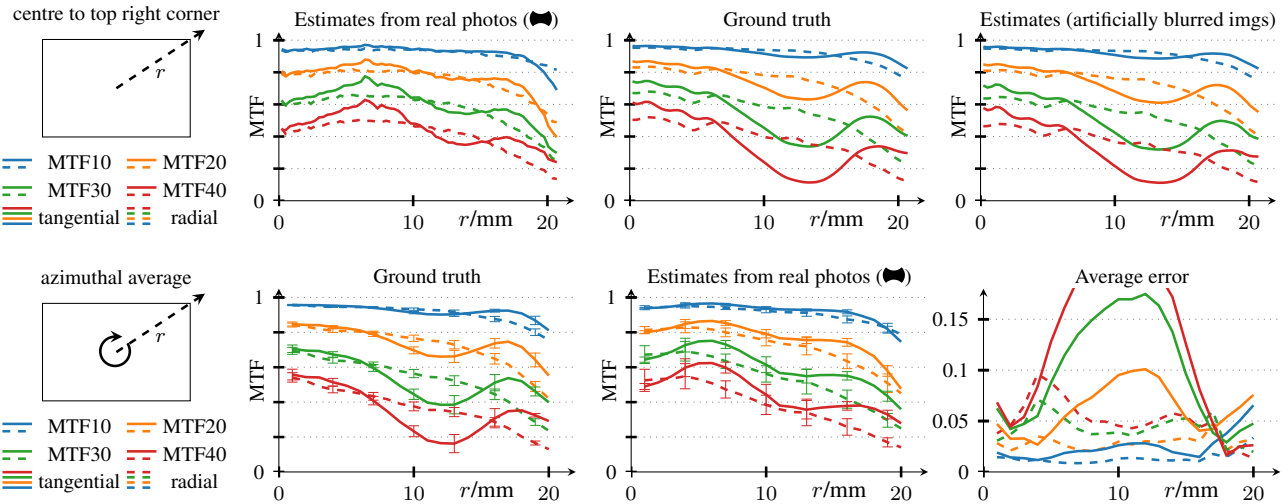

**Estimates from photos of natural scenes for Canon EF 35mm f/1.4 USM @ f/2.8**

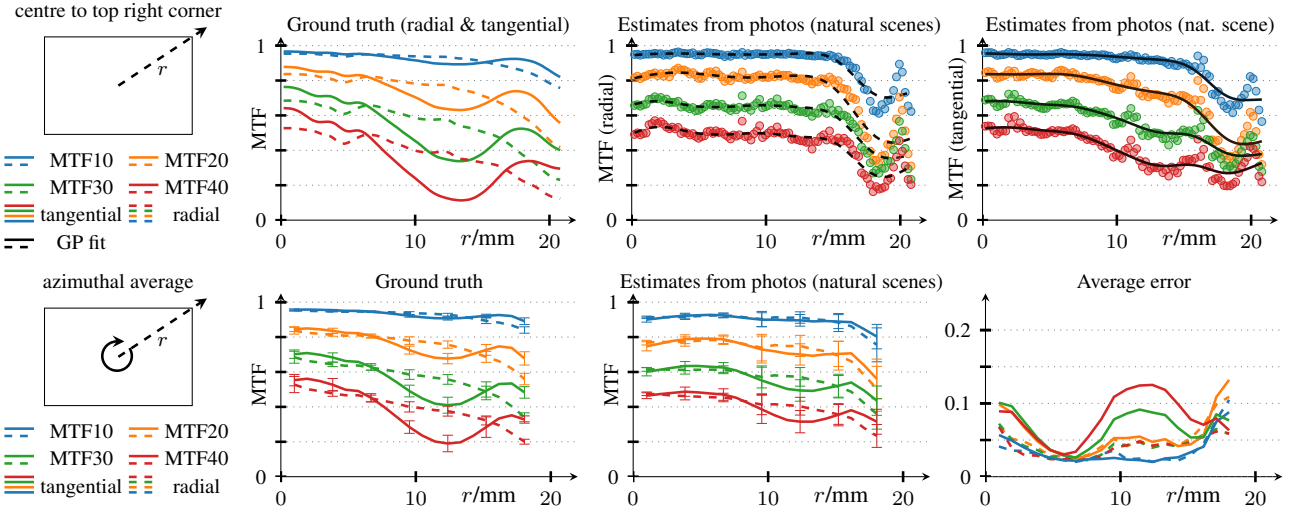

**Estimates from photos of natural scenes for Canon EF 35mm f/1.4 USM @ f/5.6**

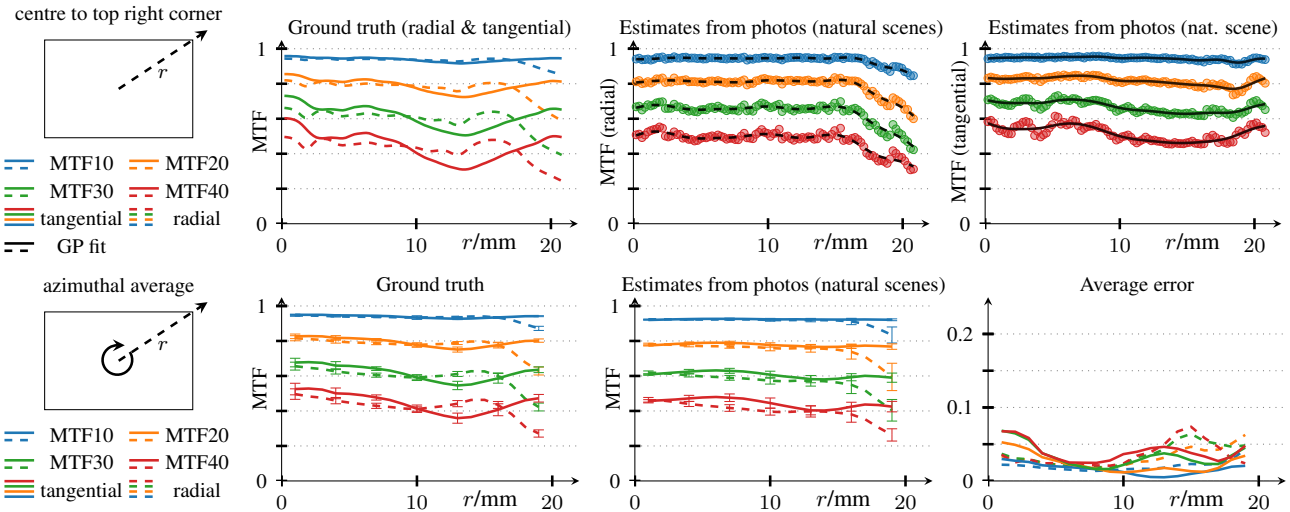

Figure 23. Results of MTF estimation for the regular pattern 🎀 (top two rows) and for natural scenes (bottom four rows) for the Canon EF 35mm f/1.4 USM lens. Results are from the centre to the top left and top right corner as well as averaged over all angles (see little cartoon on the left).

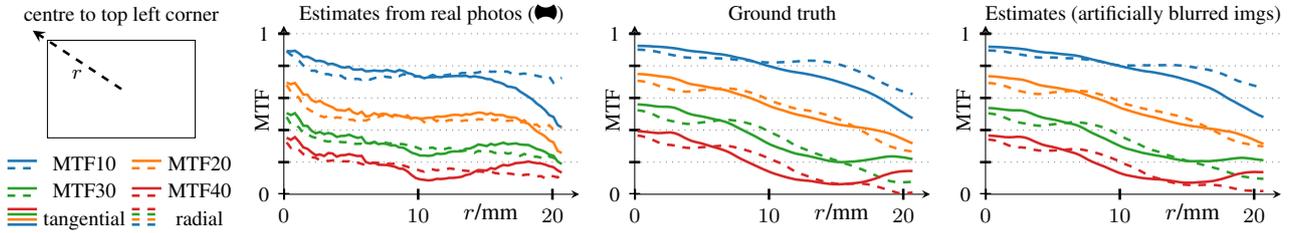
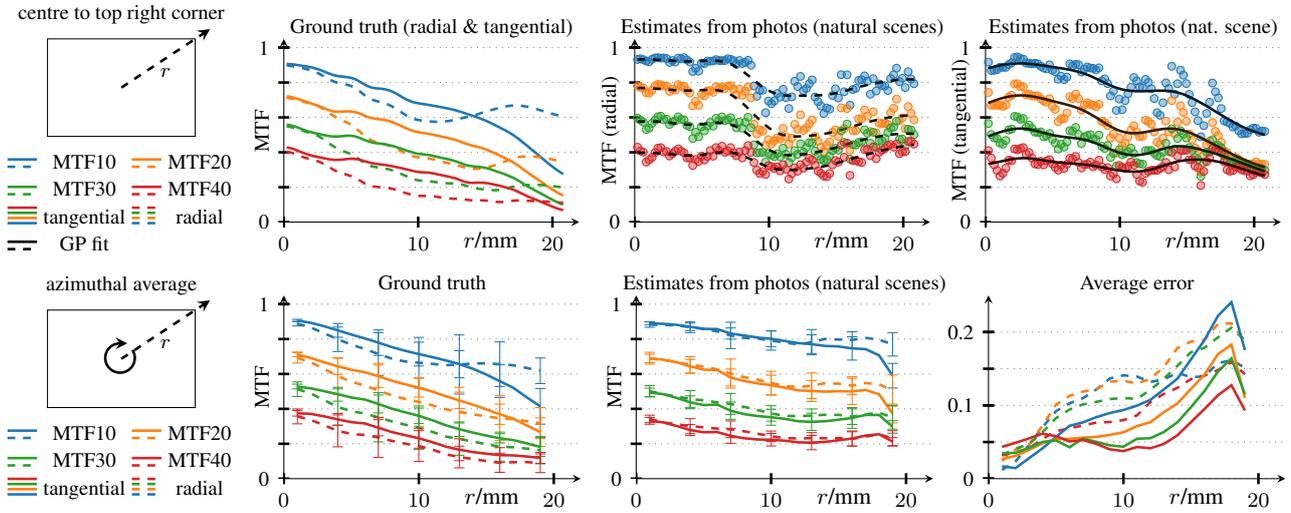
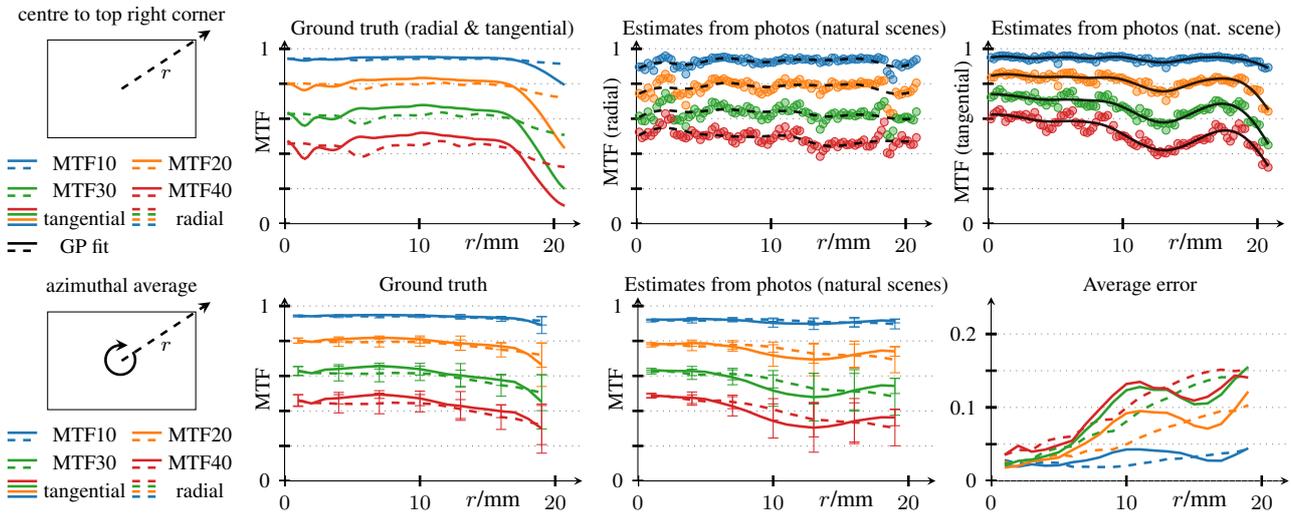

Figure 24. Results of MTF estimation for the regular pattern 🩻 (top row) and for natural scenes (bottom four rows) for the Canon EF 50mm f/1.4 USM lens. Results are from the centre to the top left/right as well as averaged over all angles (see little cartoon on the left).

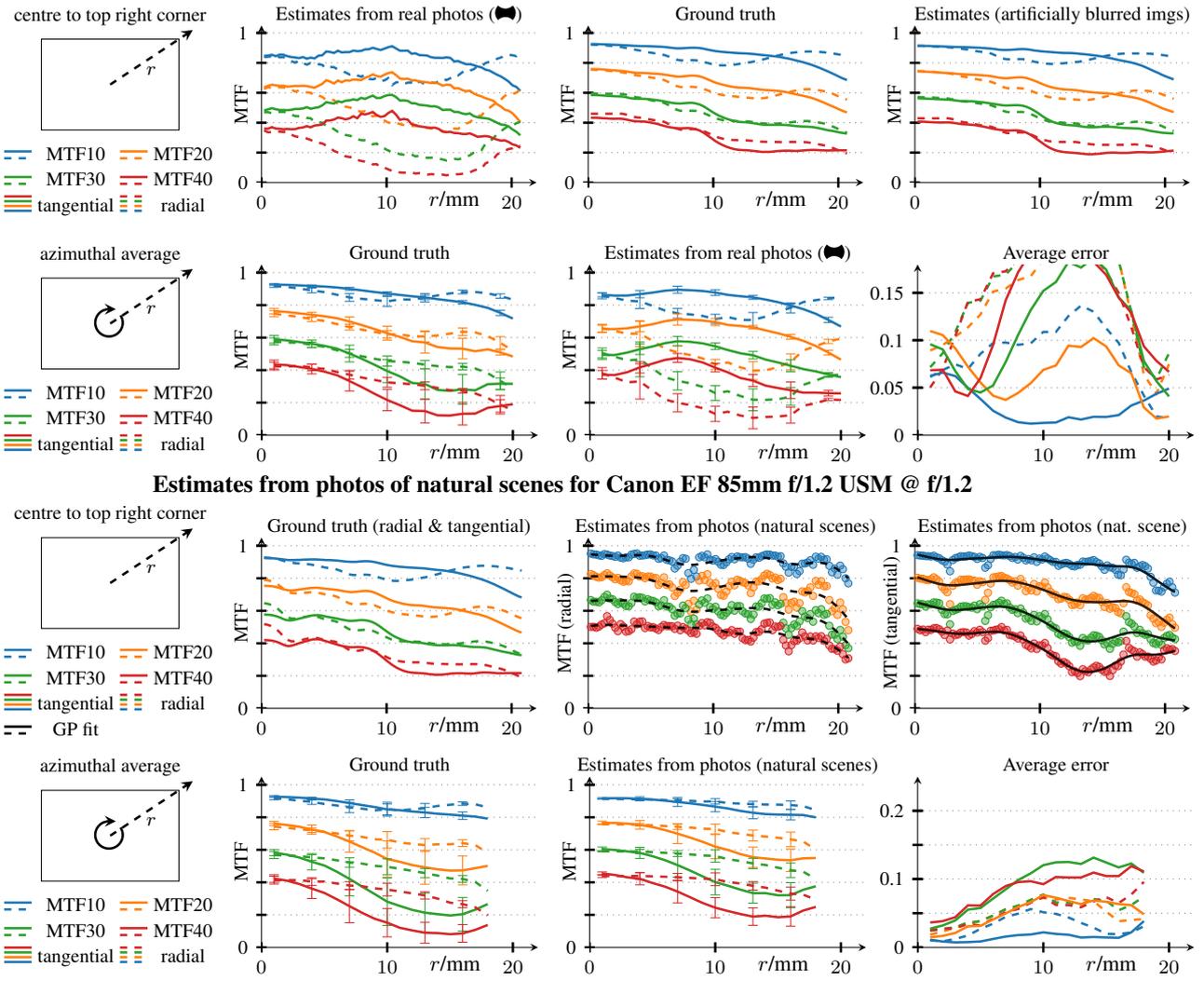

Figure 25. Results of MTF estimation for the regular pattern ▰ (top three rows) and for natural scenes (bottom two rows) for the Canon EF 85mm f/1.2 USM. Results are from the centre to the top left and top right corner as well as averaged over all angles (see little cartoon on the left).

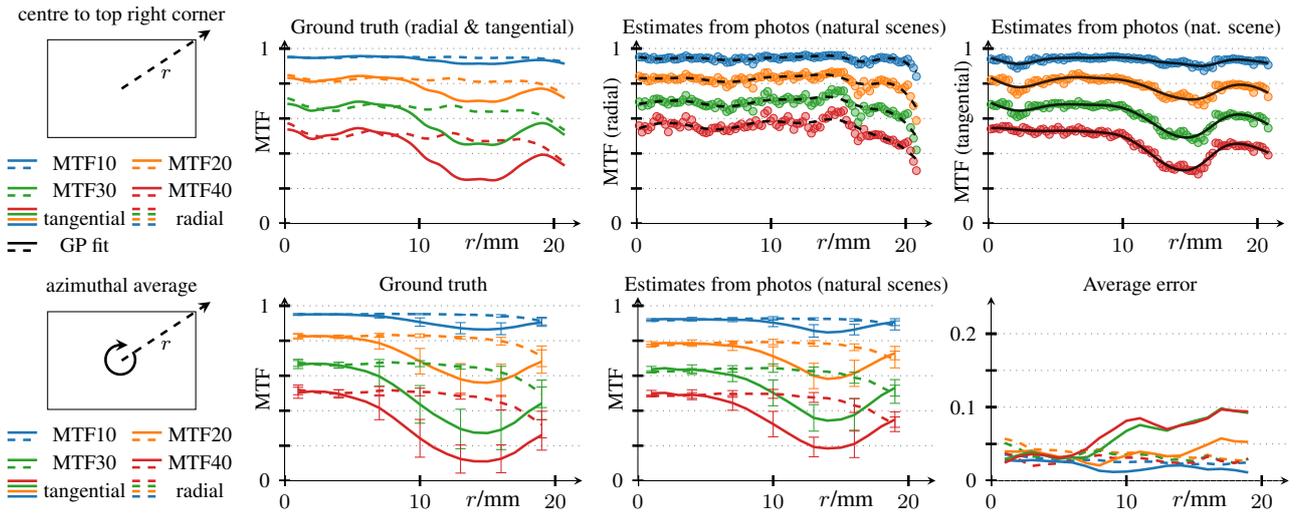

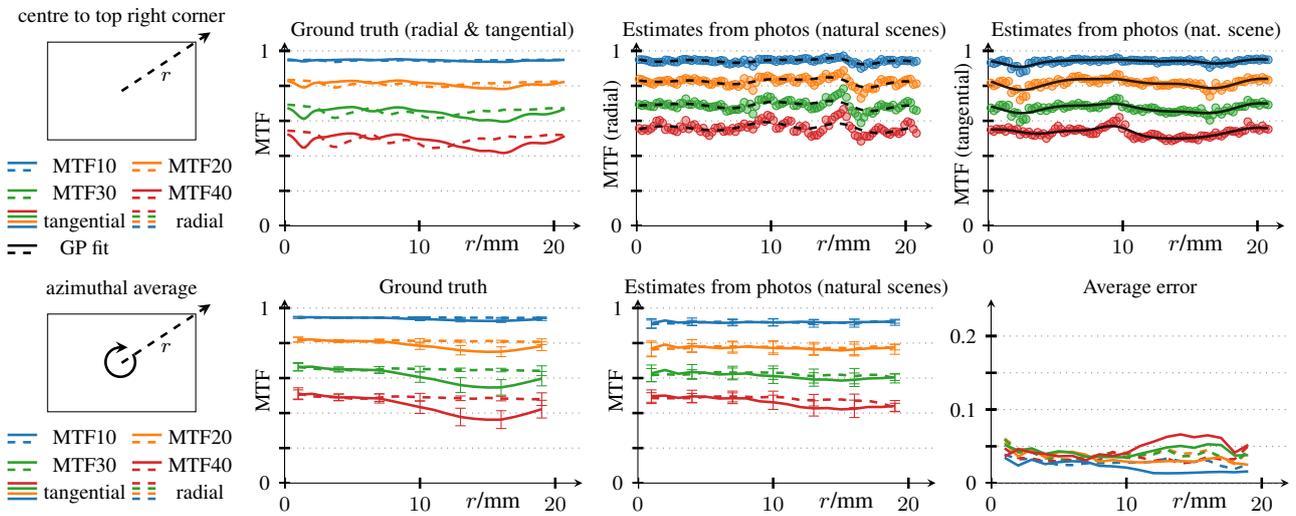

Figure 26. Results of MTF estimation from natural scenes for the Canon EF 85mm f/1.2 USM. Results are from the centre to the top right corner as well as averaged over all angles (see little cartoon on the left).

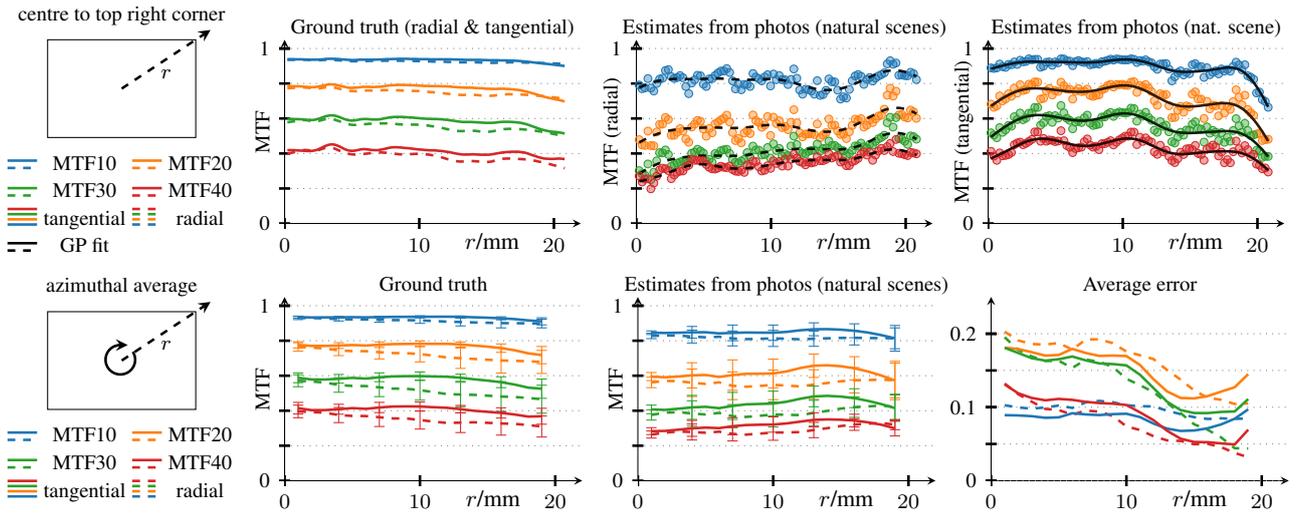

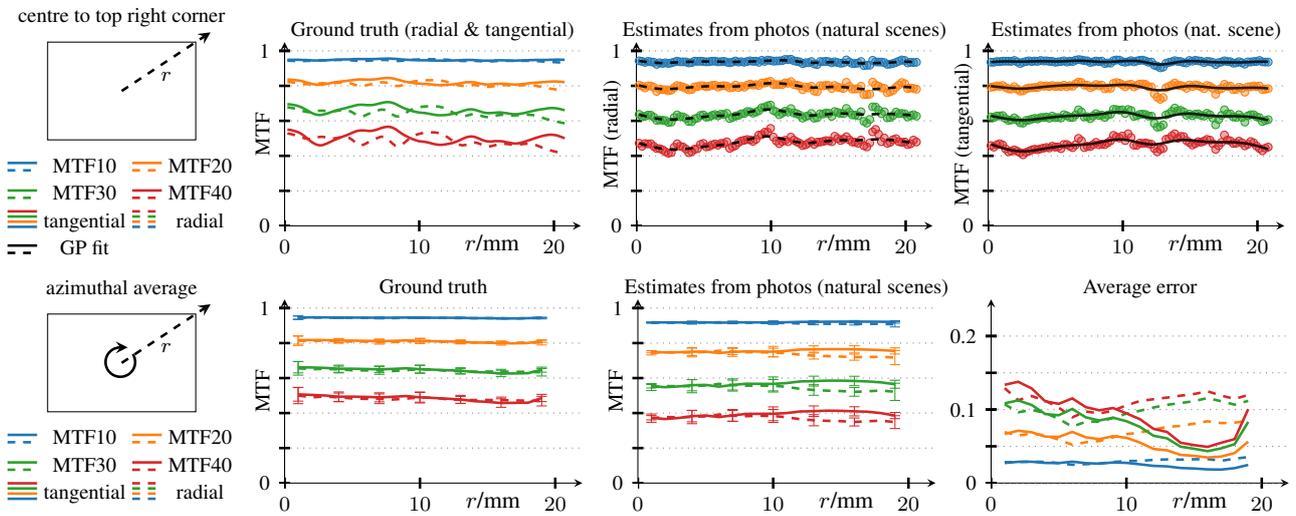

Figure 27. Results of MTF estimation from natural scenes for the Canon EF 135mm f/2.0L USM. Results are from the centre to the top right corner as well as averaged over all angles (see little cartoon on the left).